\documentclass[10pt,journal]{IEEEtran}
\usepackage{cite}

\usepackage{jabbrv}

\ifCLASSINFOpdf
  \usepackage[pdftex]{graphicx}
  \graphicspath{{./fig/}}
  \DeclareGraphicsExtensions{.pdf,.jpeg,.jpg,.png}
\else
\fi
\usepackage{amsmath}
\usepackage{amsthm}
\usepackage{amsfonts}
\usepackage{amssymb}
\usepackage{bbm}
\ifCLASSOPTIONcompsoc
 \usepackage[caption=false,font=normalsize,labelfont=sf,textfont=sf]{subfig}
\else
 \usepackage[caption=false,font=footnotesize]{subfig}
\fi
\usepackage{url}

\usepackage{color}
\usepackage{soul}
\soulregister\cite7
\soulregister\ref7
\soulregister\pageref7

\newcommand\MYhyperrefoptionsA{bookmarks=true,bookmarksnumbered=true,
pdfpagemode={UseOutlines},plainpages=false,pdfpagelabels=true,
colorlinks=true,linkcolor={black},citecolor={black},urlcolor={black},
pdftitle={Bare Demo of IEEEtran.cls for Computer Society Journals},
pdfsubject={Typesetting},
pdfauthor={Michael D. Shell},
pdfkeywords={Computer Society, IEEEtran, journal, LaTeX, paper,
             template}}

\newcommand\MYhyperrefoptionsC{bookmarks=true,bookmarksnumbered=true,
pdfpagemode={UseOutlines},plainpages=false,pdfpagelabels=true,
colorlinks=true,linkcolor={cyan},citecolor={red},urlcolor={blue}}
\ifCLASSINFOpdf
  \usepackage[\MYhyperrefoptionsC,pdftex]{hyperref}
\else
  \usepackage[\MYhyperrefoptionsA,breaklinks=true,dvips]{hyperref}
  \usepackage{breakurl}
\fi

\usepackage{multirow}
\usepackage{multicol}

\usepackage{color}
\usepackage{soul}

\DeclareRobustCommand{\hlc}[1]{{\sethlcolor{white}\hl{#1}}}
\DeclareRobustCommand{\hly}[1]{{\sethlcolor{white}\hl{#1}}}
\DeclareRobustCommand{\hlr}[1]{{\sethlcolor{white}\hl{#1}}}

\theoremstyle{definition}

\theoremstyle{remark}

\theoremstyle{plain}

\usepackage[linesnumbered,ruled]{algorithm2e}

\usepackage{booktabs}

\usepackage[utf8]{inputenc}
\usepackage[T1]{fontenc}
\usepackage{makecell}
\usepackage[table]{xcolor}

\begin{document}
%
\title{Concept Drift Detection: Dealing with Missing Values via Fuzzy Distance Estimations}
%
%
%

\author{Anjin~Liu,~\IEEEmembership{Member,~IEEE,}~Jie~Lu,~\IEEEmembership{Fellow,~IEEE,}~and~Guangquan~Zhang
}

%
%

\markboth{IEEE TRANSACTIONS ON FUZZY SYSTEM, VOL. XX, NO. X, MARCH 2020}
{Shell \MakeLowercase{\textit{et al.}}: Bare Demo of IEEEtran.cls for IEEE Journals}
%



\maketitle

\begin{abstract}
In data streams, the data distribution of arriving observations at different time points may change – a phenomenon called concept drift. While detecting concept drift is a relatively mature area of study, solutions to the uncertainty introduced by observations with missing values have only been studied in isolation. No one has yet explored whether or how these solutions might impact drift detection performance. 
\hlr{We, however, believe that data imputation methods may actually increase uncertainty in the data rather than reducing it. We also conjecture that imputation can introduce bias into the process of estimating distribution changes during drift detection, which can make it more difficult to train a learning model.} 
Our idea is to focus on estimating the distance between observations rather than estimating the missing values, and to define membership functions that allocate observations to histogram bins according to the estimation errors. Our solution comprises a novel masked distance learning (MDL) algorithm to reduce the cumulative errors caused by iteratively estimating each missing value in an observation and a fuzzy-weighted frequency (FWF) method for identifying discrepancies in the data distribution.
\hlr{The concept drift detection algorithm proposed in this paper is a singular and unified algorithm that can handle missing values, but not an imputation algorithm combined with a concept drift detection algorithm.} Experiments on both synthetic and real-world data sets demonstrate the advantages of this method and show its robustness in detecting drift in data with missing values. 
\hly{The results show that, compared to the best-performing algorithm that handles imputation and drift detection separately, MDL-FWF reduced the average drift detection difference from 10.75\% to 5.83\%. This is a nearly 46\% improvement.}
These findings reveal that missing values exert a profound impact on concept drift detection, but using fuzzy set theory to model observations can produce more reliable results than imputation.
\end{abstract}


%
\IEEEpeerreviewmaketitle

%
%
%
%






\section{Introduction}

In life, ideas constantly change and evolve. In data streams, these evolving ideas are reflected as changes in the data distribution of arriving observations. Areas where monitoring, control, and security are important are particularly dependent on fast and accurate data distribution change detection \cite{Liu:Survey, Pratama:inctype2} – for example, mobile tracking systems that monitor user behavior change, intrusion detection systems that look for unusual activity
\cite{Zliobaite:ApplicationBook, lu2020data}. The change of distribution in data streams is referred to concept drift in the literature \cite{Gama:survey, Polikar:Survey}. 

Drift detection systems typically infer changes in a situation by compiling sets of observations at different time points, estimating the discrepancies in the data distributions for each set, and comparing the results \cite{Liu:Survey}. 
For machine learning models, the inevitability of concept drift means keeping up-to-date and adapting to the current state of ideas through incremental training as new observations arrive \cite{Lu:AI1, Lu:AI2}. Without these continual self-refinements to the model, classification accuracy will eventually degrade beyond usefulness \cite{Pratama:Evolving, Pratama:pClass, dhika:muse, yiliao:TFS}. 

However, like learning with streaming data, a common problem in exploratory data analysis is handling missing values \cite{TFS:FaultDetection2017, TFS:TrackingControl2017}. Missing values in data are extremely common for dozens of reasons – human error, faulty sensors, power outages, etc. – and can cause serious implications for machine learning \cite{MissingSurvey2019}. One of the most serious of these implications is the increased levels of uncertainty an incomplete data set brings \cite{StatisticalMissing2019}. This makes concept drift detection a much more complex task. 

A common practice to mitigate the problem is to apply an imputation method to fill in the missing values with plausible substitutes \cite{MissingSurvey2019}. 
\hlc{
As discussed in \cite{GAIN:Imput}, the state-of-the-art imputation methods fall into two major categories: discriminative and generative. Discriminative methods include MICE \cite{MICE}, MissForest \cite{MissForest}, and matrix completion \mbox{\cite{Mazumder:Spectral, Yu:NIPS, Schnabel:ICML}}. Generative methods include algorithms based on either expectation maximization \cite{Miss:Review} or deep learning \mbox{\cite{Vincent:ICML, MIDAS:Imput, GAIN:Imput}}. Regardless of the category, these algorithms were proposed to resolve missing values under different circumstances. These imputation algorithms have achieved remarkable success and can be easily combined with any learning algorithms. However, most are based on value-oriented imputation, which means they focus on minimizing the difference between the estimated values and the ground truths. Yet, the flexibility is a double-edged sword, they do not care about what the data is going to be used for. In other words, they do not take a task-oriented perspective.}

\begin{figure*}[ht]
    \centering
    \includegraphics[scale=1]{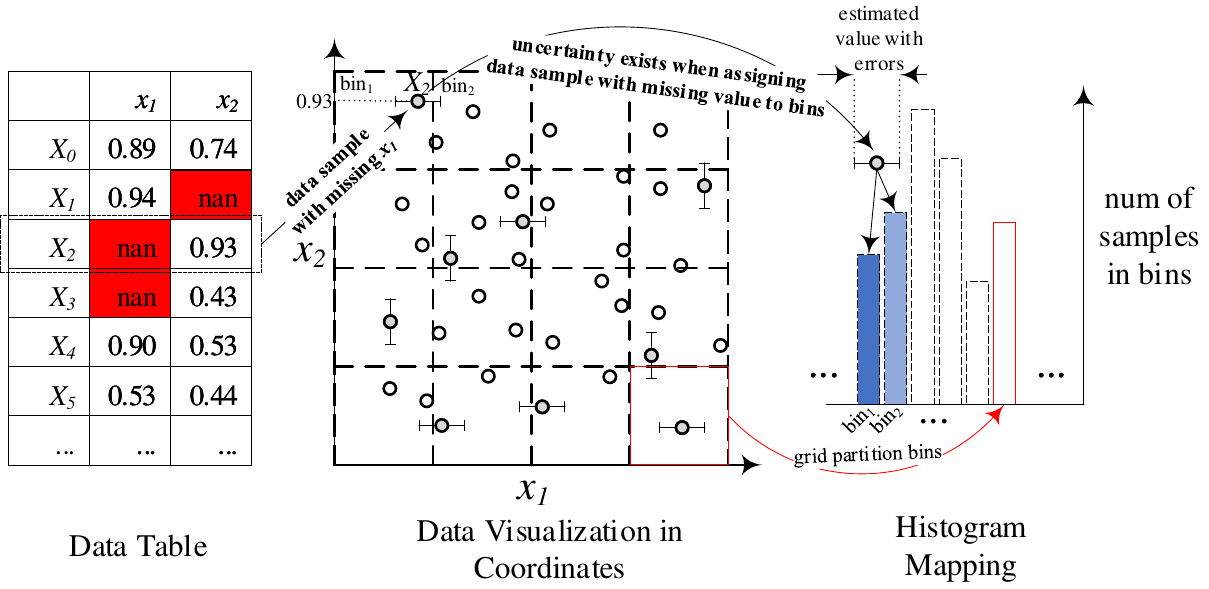}
    \caption{
    A demonstration of uncertainty when allocating data observations with missing values to histogram bins. The grey points are the samples with missing values. |-o-| indicates the confidence interval of the imputation result. The $nan$ stands for not a number. In this example, we want to use a histogram to estimate the distribution of the data set. Without considering the confidence interval, the top-left grey sample will be counted in $\mathrm{bin_1}$. Assigning $X_2$ to $\mathrm{bin_1}$ may introduce bias because there is a small chance that $X_2$ belongs to $\mathrm{bin_2}$. Therefore, we suggest splitting this sample and counting it in both $\mathrm{bin_1}$ and $\mathrm{bin_2}$ where the fuzzy set theory is the best tool for this task.
    }
    \label{fig:1}
\end{figure*}

Similarly, most drift detection researchers consider imputation to be an independent research task and so intentionally leave mechanisms for handling missing values out of their frameworks. We, however, believe that data imputation methods can have a significant impact on the accuracy and robustness of subsequent drift detection processes. 
An inappropriate value estimation will increase the bias of the statistics when comparing distribution, resulting in increased errors, such as shown in Fig. \ref{fig:1}. \hlc{Therefore, we propose a new concept, namely task-oriented imputation, that is, we directly minimize the difference between the ground truth of the statistics we need and the statistics acquired by imputation. In our solution, the statistics are the distances between observations.}


The fundamental reason that data imputation impairs the performance of a drift detection method is because the replaced values decrease data variance (\textbf{Problem 1}). And, further, errors with the missing value estimations are not considered in the distribution estimations (\textbf{Problem 2}). For example, using the mean of data as a replacement for the missing values forces the data toward a central focal point and reduces sample diversity \cite{Allison:MissingData, AppliedMissingData2010}. 
Small errors aggregate into large errors, introducing bias into the hypothesis testing for concept drift detection. Meanwhile, considering data imputation and concept drift detection as two independent tasks subjectively ignores the likelihood between the replaced value and the true value. For instance, assume we have a set of 1-dimensional observations
$\{X_i\}_{i=1}^m$, where $m$ is the size, and among these observations, there are $z$ missing values $\{X_i^{miss}\}_{i=1}^z$. The imputation is the mean of the non-missing observations denoted as $\mu$. Then the error that $\epsilon=\sum_{i=1}^z|X_i^{miss}-\mu|$
would not be considered in concept drift detection.
\hlr{
Uncertainty in data and its impact on learning models has been well studied. For instance, in \cite{TCYB_Uncertainty, TFS_Uncertainty}, the authors evaluate the performance of classifiers in terms of their fuzziness and the problem's complexity, and discover that the generalization ability of a classifier is closely related to both its fuzziness and the target problem. Extrapolating on this conclusion, we suspect that the accuracy of a drift detection algorithm may also be closely related to data uncertainty and the fuzziness of the drift detection methodology.}

To address Problem 1, we proposed a masked distance learning algorithm. The idea is to convert the distances calculated by different data imputation algorithms into a new feature set, and considering the new feature set and the true distance as a regression learning task, thereby, minimizing the difference between the estimated distance and the true distance, keeping the variance of data. 
To address Problem 2, the estimated distances are considered to be fuzzy to account for possible errors. This is coupled with a novel weighted frequency measuring method that tallies observations based on their degree of membership to histogram bins. Membership considers estimation errors in the data imputation process. 

To detect drifts, we apply Pearson’s chi-square test on the observed frequency as a hypothesis test \cite{box:sampleSize, Liu:TCYB2020}, and the significance level is governed by a parameter $\alpha$,  which determines the sensitiveness of the hypothesis test to concept drift.
In summary, the main contributions of this paper are:
\begin{itemize}

\item The notion of estimating the distance between samples rather than estimating the missing values, which reduces the cumulative errors caused by iterative estimations.

\item A novel concept drift detection algorithm that is robust to missing values, along with a masked distance learning algorithm and a fuzzy-weighted frequency observing method – MLD-FWF when combined.

\item A ablation study was conducted to evaluate the improvement of using fuzzy set theory and not using fuzzy set theory. The results indicate that applying fuzzy set theory for drift detection with missing values can achieve more reliable result.

\item A comprehensive evaluation of the proposed algorithms and framework on both synthetic and real-world data sets with results revealing that applying fuzzy set theory to handle missing values is beneficial to reduce false alarms in concept drift detection.
\end{itemize}


The rest of this paper is organized as follows. In Section \ref{s:II}, we discuss the problem of concept drift detection and data imputation, followed by the preliminaries relating to our drift detection method. Section \ref{s:III} presents the proposed masked distance learning and the fuzzy-weighted frequency drift detection algorithms. Section \ref{s:IV} evaluates the proposed distance learning performance and drift detection accuracy. Section \ref{s:V} concludes this study with a discussion of future work.

\section{Literature Review and Preliminaries}
\label{s:II}
\hly{This section presents the problem of concept drift detection (Section \ref{s:II.A}) and missing values (Section \ref{ss:mv}). The preliminaries of our solution - Pearson's chi-square test is also introduced, Section \ref{ss:Chi2}}
\subsection{Concept Drift Detection}
\label{s:II.A}

In an evolving data stream, the distribution of available training samples may vary over time \cite{Liu:Survey}. Consider a topological space feature space denoted as $\mathcal{X}\subseteq \mathbb{R}^n$, where $n$ is the dimensionality of the feature space. A tuple $(X,y)$ denotes a data instance, where $X\in\mathcal{X}$ is the feature vector, $y\in \{y_1,\ldots,y_c \}$ is the class label, and $c$ is the number of classes. A data stream, denoted as $\mathcal{D}$, can then be represented as a sequence of data instances. Time windows strategy chunks the sequence in a time interval as a batch. These batches are denoted as $D_{T_i}\in\mathcal{D}$, where $T_i$ is a given time interval that defines the time window. A concept drift has occurred between two time windows $T_i$ and $T_{i+1}$ if the joint probability of $X$ and $y$ is different, that is, $p_{T_i}(X,y)\neq p_{T_{i+1}}(X,y)$ \cite{Gama:survey, Polikar:Survey}.

There are two approaches to estimating the density discrepancy: parametric and nonparametric \cite{Silverman:2018, Liu:TCYB2020}. The parametric approach assumes the data is drawn from a known distribution, with the major challenge being how to estimate the parameters to reach the maximum likelihood \cite{Liu:TCYB2020}. Nonparametric approaches assume that it is too difficult to devise an analytic expression of the target distribution's probability density function \cite{Alippi:LSDD-CDT}. Therefore, the density is estimated empirically rather than by matching the data with any given parametric family. Given that concept drift detection inherently concerns evolving data distributions, it is difficult to rely on one distribution family for an accurate description. Therefore, most drift detection methods are nonparametric \cite{Liu:Survey}.

To quantify the distribution discrepancy between two multivariate sample sets, some researchers have proposed new test statistics, while others have proposed novel mapping methodologies that convert multivariate observations into univariate observations. For example, NN-DVI \cite{Liu:PR} and MMD \cite{Gretton:MMD} construct new statistical variables for comparing distributions to determine the differences. The performance of these methods is excellent, but that performance usually comes with a high computational cost. Alternatively, if the distribution of the proposed statistics is unclear, a Monto Carlo method can be applied to approximate the significance level. Examples include kdqTree combines with Kullback-Leibler divergence \cite{Dasu:kdqTree}, competence model combined with total variation \cite{Lu:AI1}, QuantTree combined with total variation, or QuantTree combined with Chi-square statistics \cite{Boracchi:QTree}. On yet another track, the solution involves building a model to map the discrepancy in distributions with powerful statistical tools, such as Pearson’s Chi-square \cite{Liu:TCYB2020}. The general premise of these approaches is to convert multivariable data samples into a univariable two-sample test problem, such as an incremental Kolmogorov–Smirnov test \cite{FastKSTest} or a multi-Wald-Wolfowitz test \cite{MWWtest}. However, no research has considered the missing values for concept drift detection.

\subsection{Types of Missing Values and Handling Methodologies}
\label{ss:mv}
Missing values are merely observations that were not recorded, perhaps due to human error or a sensor malfunction, which is very common in real-world data \cite{Honda:fuzzyCluster,StatisticalMissing2019,MissingSurvey2019}. There are three main categories of missing values based on the correlation between the observations and the values that are missing. These are: missing completely at random (MCAR), missing at random (MAR) and missing not at random (MNAR) \cite{AppliedMissingData2010,FlexibleImputation2018}. A recent survey by  Santos et al. \cite{MissingSurvey2019} provides a set of notations and definitions that demonstrate the difference between each category quite well, as follows:

Assume we have a data set with $m$ number of instances and $n$ number of features. The values of the data set are represented as an $m\times n$ matrix as $X=\{x_{i,j}\}_{i,j}^{m,n}$. Plus, a missing data indicator matrix is defined as an $m\times n$ zero-one matrix, denoted as $M$. $M$ is represented as $M=\{I_{i,j}^{miss}\}_{i,j}^{m,n}$, where $I_{i,j}^{miss}$ is the missing value indicator and equals 1 if $x_{i,j}$ is missing, and 0 otherwise. The relationship describing the missing data is defined as a conditional probability function $p(M|X,\xi)$, where $\xi$ are the parameters of the missing values \cite{Allison:MissingData}. In practice, $\xi$ is not important, only the relation between $M$ and the components of $X$ are important, i.e., $X=(X^{complete},X^{miss} )$, where $X^{complete}$ is the observations without missing values, and $X^{miss}$ is the the observations with missing values.

MCAR means that a missing value has nothing to do with its hypothetical value or with the values of other variables \cite{MissingSurvey2019,StatisticalMissing2019}. The probability of missingness depends only on the parameters $\xi$, i.e., $p(M=1|X,\xi)=p(M=1|\xi)$ \cite{MissingSurvey2019}.

MAR means the missing values are not related to themselves, but they do depend on other features somehow. In other words, the conditional probability of missing values is $p(M=1|X,\xi)=p(M=1|X^{complete},\xi)$ [5].

With MNAR, the missing values depend on both the observed and the unobserved information – $X^{complete}$ and $X^{miss}$ – and the conditional probability of the missing values cannot be simplified, i.e., $p(M=1|X,\xi)=p(M=1|X^{complete},X^{miss},\xi)$ \cite{MissingSurvey2019}.
Conventional methods of handling missing values include complete case analysis, available case analysis, dummy variable adjustment, and imputation \cite{Allison:MissingData}.




Imputation is currently the most widely used method for handling missing values \cite{StatisticalMissing2019}. Like dummy variable adjustment, missing values are replaced but this time with plausible values, not constants or a mean. The model is then trained just as though no values were missing \cite{Allison:MissingData}. However, there are two serious problems with most imputation methods. First, variances tend to be underestimated, which can lead to bias in the learning parameters \cite{Allison:MissingData}. Second, the standard error calculations presume that all data are real, which means the inherent uncertainty and sampling variability in the imputed values is not taken into account \cite{Allison:MissingData}. Of course, this results in confidence intervals that are too narrow.

\subsection{Pearson’s Chi-square Test}
\label{ss:Chi2}
Pearson’s Chi-square Test has been widely used to determine whether there are significant differences between expected frequencies and observed frequencies \cite{box:sampleSize}. When there is no significant difference (i.e., when the null hypothesis is true), the statistic follows a chi-square distribution. Thus, the premise of the test is to assume the null hypothesis is true and then evaluate how likely a specific observation would be.

The standard process of the chi-square test is to use sample data to find: the degrees of freedom, the expected frequencies, the test statistic, and the p-value associated with the test statistic \cite{box:sampleSize}. Given a contingency table with $i$ rows and $j$ columns of categorical variables, the degrees of freedom are equal to 
\begin{equation*}
    DF = (i-1)(j-1).
\end{equation*}
The expected frequency counts are computed separately for each level of one categorical variable at each level of the other categorical variable. The $i$-th and $j$-th expected frequencies of the contingency table are calculated with the equation 
\begin{equation*}
    E_{i,j} = (n_i\times n_j)/n,
\end{equation*}
where $n_i$ is the sum of the frequencies for all columns in row $i$, $n_j$ is the sum of the frequencies for all rows of columns $j$, and $n$ is the sum of all rows and columns. The test statistic is a chi-square random variable $\chi^2$ defined by 
\begin{equation}
\label{equ:chi2}
    \chi^2 = \sum \frac{(O_{i,j}-E_{i,j})^2}{E_{i,j}},
\end{equation}
where $O_{i,j}$ is the observed frequency count at row $i$ and column $j$, and $E_{i,j}$ is the expected frequency count at row $i$ and column $j$. The p-value is the probability of observing a sample statistic as extreme as the test statistic. Since the p-value is a $\chi^2$ test statistic, it can be computed with the chi-square probability distribution function.

\hlc{Pearson's chi-square test should be used with a sufficiently large sample set. According to the central limit theorem, an $\chi^2$ distribution is the sum of $O_{i,j}$ independent random variables with a finite mean and variance that converges to a normal distribution for large $O_{i,j}$. For practical purposes, Box et al. \cite{box:sampleSize} claim that, for $O_{i,j}>50$ and $E_{i,j}>5$, the distribution of the estimated test statistics is sufficiently close to a normal distribution to ignore the difference.}
\section{Concept Drift Detection with Missing Values via Fuzzy Distance Estimations}
\label{s:III}
\hly{This section formally presents our proposed solution for concept drift with missing values. In Section \ref{ss:dd_via_hist}, we present a histogram-based concept drift detection method. Section \ref{ss:mdl} introduces our strategy for handling missing values. Last, Section \ref{ss:fwf} discusses how we leverage fuzzy distance estimations to address the uncertainty of missing values in drift detection.}


\subsection{Concept Drift Detection via Histogram Distribution Estimation}
\label{ss:dd_via_hist}

\textbf{Problem Statement}. Recall the drift detection problem outlined in Section \ref{s:II.A}, i.e., Let $\mathcal{A}$ and $\mathcal{B}$ be random variables defined on a topological space $\mathcal{X}\subseteq \mathbb{R}^n$ at different periods in a data stream with respect to $p_{\mathcal{A}},p_{\mathcal{B}}\in \mathcal{P}(\mathcal{X})$, where $\mathcal{P}(\mathcal{X})$ consists of all Borel probability measures on $\mathcal{X}$. Given two observation sets drawn from $\mathcal{A}$ and $\mathcal{B}$,  $A=\{X_i^\mathcal{A}\}_{i=1}^{m_1}$ and $B=\{X_i^\mathcal{B}\}_{i=1}^{m_2}$, the problem is: How confident are we to claim that $\mathcal{A}=\mathcal{B}$ based on $A$ and $B$? 

To simplify this problem, it is commonly assumed that the observations $A$, $B$ are i.i.d., which makes the research objective equivalent to a two-sample test problem. The general procedure is to measure the difference between the observations $A$ and $B$ (sample sets), then infer the significance of the difference between $\mathcal{A}$ and $\mathcal{B}$ (population). 

\textbf{Remark}: Recall the definition of concept drift, $p_{T_i}(X,y)\neq p_{T_{i+1}}(X,y)$, a data instance is defined as a tuple of ($X$, $y$). For supervised learning setting, the target variable $y$ can be considered as the $(n+1)$-th feature of the observations while for unsupervised setting, the drift detection will be equivalent to covariate shift detection \cite{Jose:ConceptShift}. 

\begin{figure}[ht]
    \centering
    \includegraphics[scale=0.5]{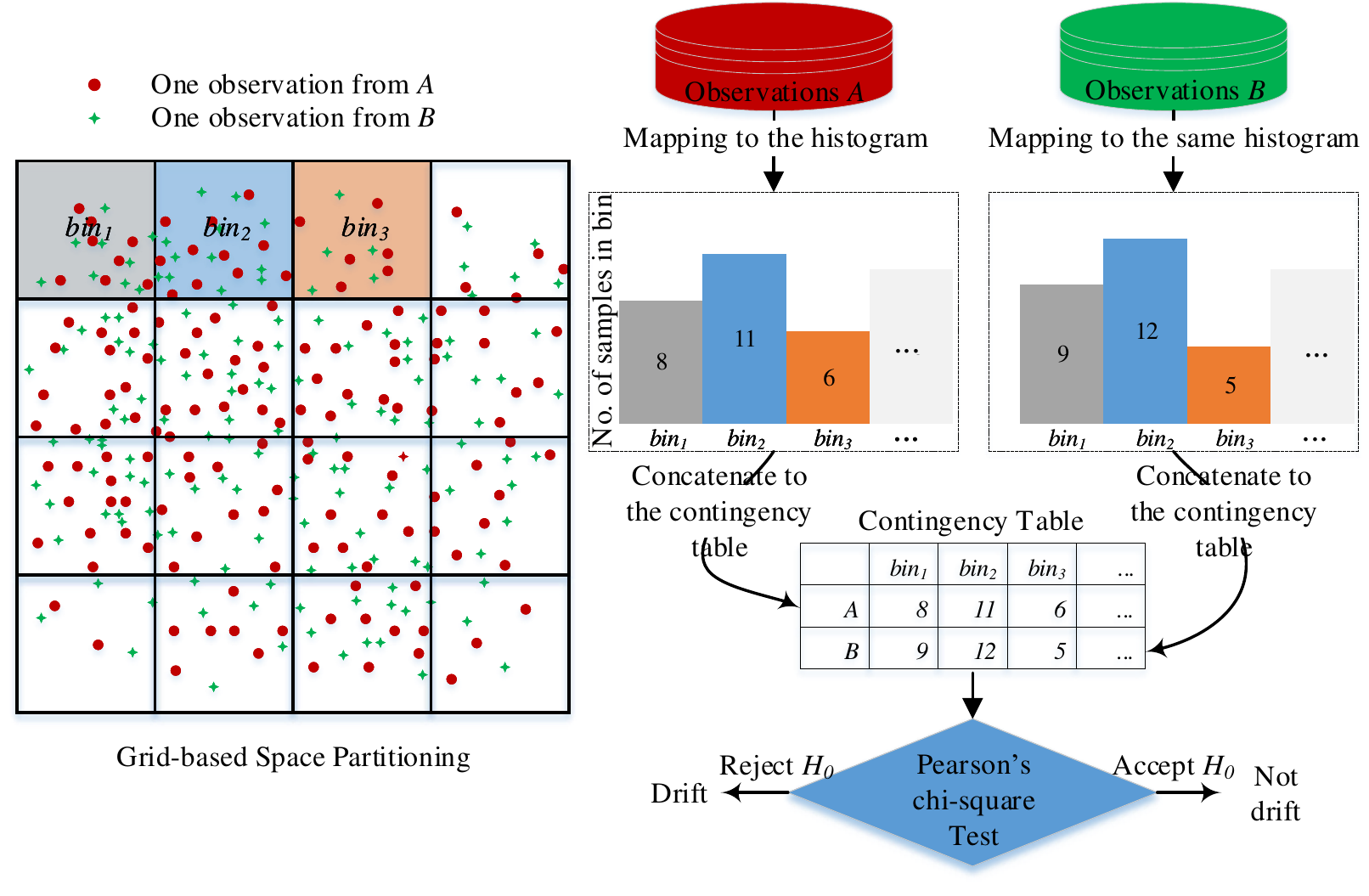}
    \caption{
    The process of histogram-based concept drift detection. The number of observations in each partition are counted and then transformed into a contingency table. Pearson’s chi-square test outputs the significance of the difference, which is not drift below a threshold and drift above. Note that this diagram is for demonstration purposes only; the number of observations in each bin would need to be much higher (e.g., 50 per bin minimum) for the chi-square approximation to be valid \cite{box:sampleSize}.
    }
    \label{fig:2}
\end{figure}

The most intuitive method for capturing this difference is a histogram. A histogram is a set of intervals, i.e., bins, and density is estimated by counting the number of observations in each bin. A general framework for using a histogram and Pearson’s chi-square test to detect concept drift is shown in Fig. \ref{fig:2}.

\begin{figure*}[ht]
    \centering
    \includegraphics[scale=0.43]{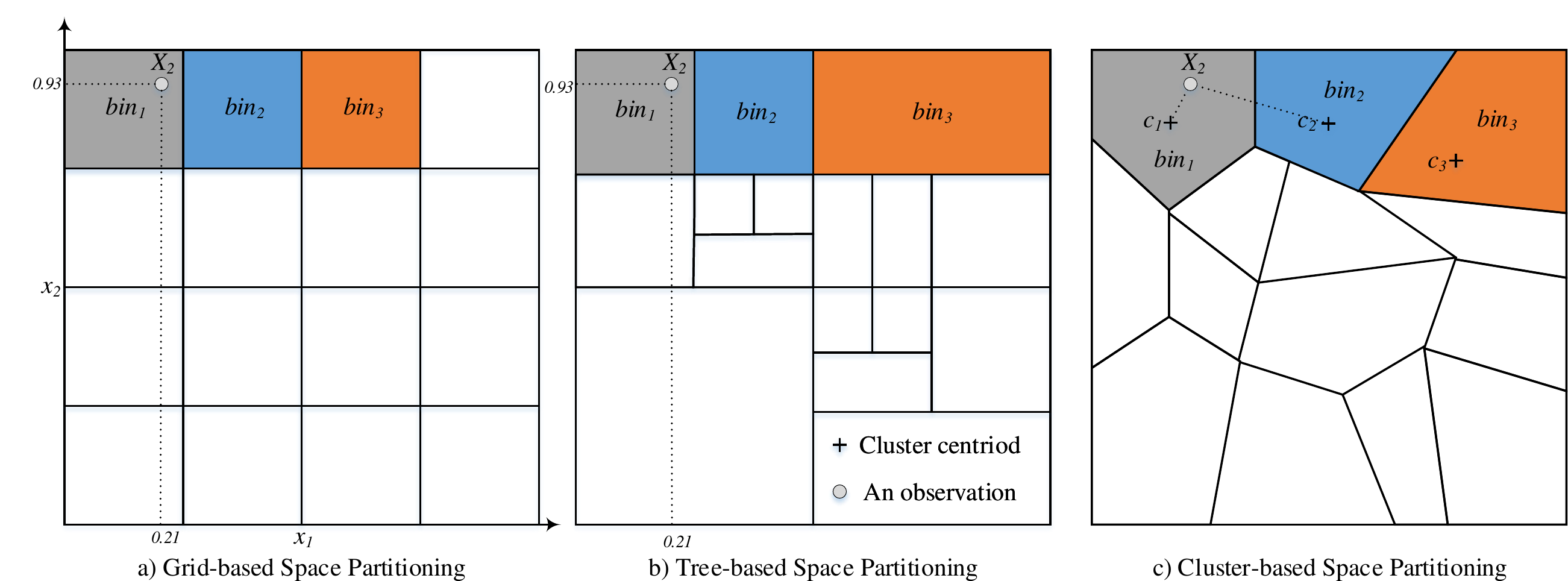}
    \caption{
    A demonstration of different space partitioning algorithms: Consider an observation $X=(x_1=0.21,x_2=0.93)$. Grid- and tree-based algorithms match the value of the features to the bin intervals, whereas cluster-based algorithms compare the observation’s distance to each centroid.
    }
    \label{fig:3}
\end{figure*}

The next question is how to partition the feature space as bins to build a histogram. In the literature, the partitions of most multivariate histograms are based on a grid or a tree structure \cite{Boracchi:QTree,Liu:TCYB2020}. We have used clusters because, given our problem, this method makes it easy to determine which bin an observation belongs to. More specifically, with grid- and tree-based histograms, observations are allocated to a bin according to the value of each feature, whereas, with cluster-based histograms, the bin is determined by the distance between the observations and the centroids, as demonstrated in Fig. \ref{fig:3}.

\hlc{In practice, there are three advantages to using clusters as histogram bins.}

1) The shape of the bins is flexible. Both grid and tree-based histograms have hyper-rectangular-shaped bins. However, the shape of the bins in a cluster-based histogram dynamically adapts to the data distribution, which is beneficial when trying to maintain a similar number of observations per bin and not leaving any empty bins.

2) Missing values in an observation do not need to be estimated one-by-one with cluster-based histograms. Rather, only the observation-to-centroid distances need to be modeled to determine the appropriate observation-to-cluster membership. In addition, the error of the distance estimation can be characterized by fuzzifying the distance then using that fuzzy distance to calculate the observation’s degree of membership to a cluster.

\hlc{3) Since our algorithm is task-oriented imputation rather than value-oriented, our drift detection algorithm has no assumption on the data type, i.e., both numerical and categorical data can be addressed.
} 


\subsection{Masked Distance Learning}
\label{ss:mdl}
This section sets out our formulation for a distance learning method to estimate observation-to-observation distances with missing values. The intuition behind the idea is to retrieve the observations without missing values, denoted as $X^{complete}$, and artificially introduce null values as a mask to simulate the missing values, that is, 
\begin{equation*}
    X^{complete} \xrightarrow[]{f_{mask}(X)}  X^{mask},
\end{equation*}
where $f_{mask}(X)$ denotes the missing value masking function. Since our target variable $d$ is known, i.e., the actual paired distances between the retrieved observations, we can use the masked samples $X^{mask}$ and their actual distances $d$ as a training set to build a regression model, denoted as $L_{mask}$. The standard deviation $\sigma_{mask}$ of $L_{mask}$ can be used as a coefficient for fuzzifying observation-to-observation distances.
\begin{figure*}[ht]
    \centering
    \includegraphics[scale=0.6]{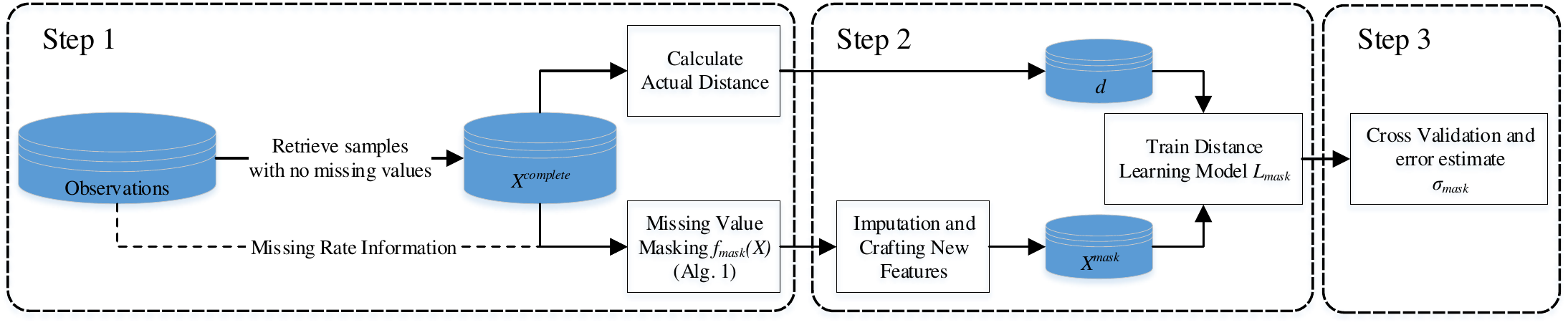}
    \caption{
    The workflow of fuzzy distance learning by masking missing values. Step 1. Define the masking function, $f_{mask}(X)$. Step 2. Craft the features and train the distance learning model $L_{mask}$. Step 3. Estimate the distance prediction error, such as the standard deviation of the residual $\sigma_{mask}$.
    }
    \label{fig:4}
\end{figure*}

The workflow of this masked distance learning method comprises three major steps, as shown in Fig. \ref{fig:4}. Put simply, masked distance learning asks and answers three questions: 1) How must the function $f_{mask} (X)$ be defined so that the observations with missing values have the same distribution as the original sample set? 2) Which method of building the learning model $L_{mask}$  will minimize the prediction errors? 3) How should the error of the $L_{mask}$ be estimated? Step 1-3 are proposed to address these questions.

Step 1 is a sample selection process. Retrieving observations with no missing values can be complete by a for loop over the sample set. The runtime complexity is $\mathcal{O}(mn)$, where $m$ is the number of observations, and $n$ is the dimension of the feature space. The number of observations with missing values is then counted and divided by the total number of observations to produce a missing rate vector, denoted as $V_r=\{r_i\}_{i=1}^n$, where $r_i$ is the missing rate of the $i$-th feature. At this stage, we assume the missing values are MCAR. Based on this assumption, the probability mass function of the masking variable for each feature is
\begin{equation}
\label{equ:1}
    P(x_i^{mask})=
    \begin{cases}
      r_i, & if ~  x_i^{mask} ~ is ~ null\\
      1-r_i, & if ~ x_i^{mask} = x_i
    \end{cases},
\end{equation}
and we have an masked observations as $X_1^{mask}=\{x_i^{mask}\}_{i=1}^n$. The actual pairwise distances of $X^{complete}$ for Step 2 is also calculated at this time.

Step 2 is the prediction model training. In this step, we want to build a learning model that maps the masked observations to the actual pairwise distance while minimizing the error between the predictions and the actual values, i.e.,
\begin{equation*}
    \mathop{\min\arg}_{h\in \mathcal{H}} |h(X^{mask})-d|,
\end{equation*}
where $\mathcal{H}$ is the hypothesis space, and $h\in \mathcal{H}$ is the hypothesis of mapping observations to a real value $h:X\rightarrow d$. Few existing data imputation methods are applied to the masked observations, and the masked distances are calculated as new training features. Then these new training features, along with the actual distances prior to masking, are used to train a masked distance learning model. \hlc{In our solution, we use the gradient boosted decision tree model to perform the learning task, which is a non-linear regression model. We consider that non-linear regression models often outperform linear models in complicated cases.}

The basic assumption is that the true distance is close to the imputed distances but which imputation method provides the best results depends on the data. For example, mean replacement is the best option for or Gaussian-distributed features, whereas most frequent fills could be the best for Poisson-distributed features. So, to make the optimal choice, the imputed distances are passed to a regression model and learned. The masked training set contains $\frac{|X^{mask}|^2}{2}$ instances for training, where $|\cdot|$ denotes the cardinality of the set. On the MCAR assumption, the masked distance learning model should perform similarly across the entire sample set.

Step 3 is to estimate the performance of the trained distance learning model $L_{mask}$. According to the assumption of linear regression, i.e., that $d_i=a+\theta X_i+\epsilon_i$ where the error item $\epsilon \sim\mathcal{N}(0,\sigma^2)$, we can use the error item $\epsilon$ (residual) to fuzzify the predicted distance. To estimate the variance, the residual standard error calculated by the equation below and then stored,
\begin{equation}
\label{equ:2}
    \sigma_{mask}=\sqrt{\frac{\sum_{i=1}^m (d_i-\hat{d}_i)^2}{m-2}}.
\end{equation}
For each predicted distance $\hat{d}_i$, we have high confidence that the true distance is located in the interval $[\hat{d}_i-3\sigma_{mask}, \hat{d}_i+3\sigma_{mask}]$ according to the three-sigma rule. To avoid sampling bias, we apply cross-validation to the $X^{mask}$, which splits $X^{mask}$ into kFold paired subsets $\{X_{train_i}^{mask},X_{test_i}^{mask}\}_{i=1}^{n_{split}}$, after which the $\sigma_{mask}$ is computed, where the $n_{split}$ is the number of split,.

Alg. \ref{alg:1}. presents the pseudocode of masked distance learning. The inputs are the observations, a default regression model, a parameter for cross-validation, and a set of data imputation methods. The outputs are a trained $L_{mask}$ and its residual standard deviation $\sigma_{mask}$. The overall runtime complexity is $\mathcal{O}(n_{split}m^2 n)$ broken down as follows. The complexity for Lines 1-3 to initialize the data set statistics is $\mathcal{O}(mn)$. The pairwise distance has a complexity of $\mathcal{O}(m^2 n)$ without any optimization. The worst case for the observation masking, i.e., when every value is missing, is $\mathcal{O}(mn)$. The worst case for the data imputation is multiple iterative imputations, which would be $\mathcal{O}(m^2 n)$. The kFold training-testing complexity is $\mathcal{O}(n_{split}m^2 n)$. Calculating standard deviations is $\mathcal{O}(m)$. Therefore, the runtime complexity for Alg. \ref{alg:1}. is feasible within the complexity of $\mathcal{O}(n_{split}m^2 n)$

\begin{algorithm}[ht]
    \caption{Masked Distance Learning (MDL)}
    \label{alg:1}
    \small
    \SetKwInOut{Input}{input}
    \SetKwInOut{Output}{output}
    \Input{1. Observations, $X$
        \newline 2. Regression model, $L_{mask}$, default GBDT regressor
        \newline 3. Cross-validation times, $n_{split}=5$ as default
        \newline 4. Data imputation algorithm pool, default $\Xi=(\xi_{zero},\xi_{mean},\xi_{med},\xi_{mfr},\xi_{iter})$
        }
    \Output{1. Trained distance learning model, $L_{mask}$
    \newline 2. Prediction error, $\sigma_{mask}$
    }
    \BlankLine
    estimate the missing rate vector $V_r$\;
    retrieve non-missing observations, $X^{complete}$\;
    calculate pairwise distances of $X^{complete}$ as $d^{mask}$\;
    \For {i $\mathrm{in}$ $\mathrm{range}$(n)} {
        \For {j $\mathrm{in}$ $\mathrm{range}$(m)} {
            mask value $x_{ji}$ according to $r_i$ \tcp*[r]{Eq. \eqref{equ:1}}
        }
    }
    
    \For {\hlc{$\xi$ $\mathrm{in}$ $\Xi$}} {
    	apply data imputation $X^{impu}=\xi(X^{mask})$\;
    	calculate pairwise distances of $X^{impu}$ as $x_{2n+i}$\;
    	concatenate $x_{2n+i}$ to $X^{mask}$\;
    }
    
    \For {$i$ $\mathrm{in}$ $\mathrm{range}$($n_{split}$)} {
        split $X^{mask}$,$d^{mask}$ into ($X_{train}^{mask}$,$d_{train}^{mask}$,$X_{test}^{mask}$,$d_{test}^{mask}$ )\;
        train $L_{mask}$ with $X_{train}^{mask}$,$d_{train}^{mask}$\;
        valid $L_{mask}$ with $X_{valid}^{mask}$,$d_{valid}^{mask}$ and store the residuals\;
    }
    compute the standard deviation $\sigma_{mask}$ \tcp*[r]{Eq. \eqref{equ:2}}    
    train $L_{mask}$ with $X^{mask}$, $d^{mask}$\;
    return ($L_{mask}$, $\sigma_{mask}$)\;
\end{algorithm}

\subsection{Fuzzy-weighted Frequency}
\label{ss:fwf}

\begin{figure}[ht]
    \centering
    \includegraphics[scale=0.6]{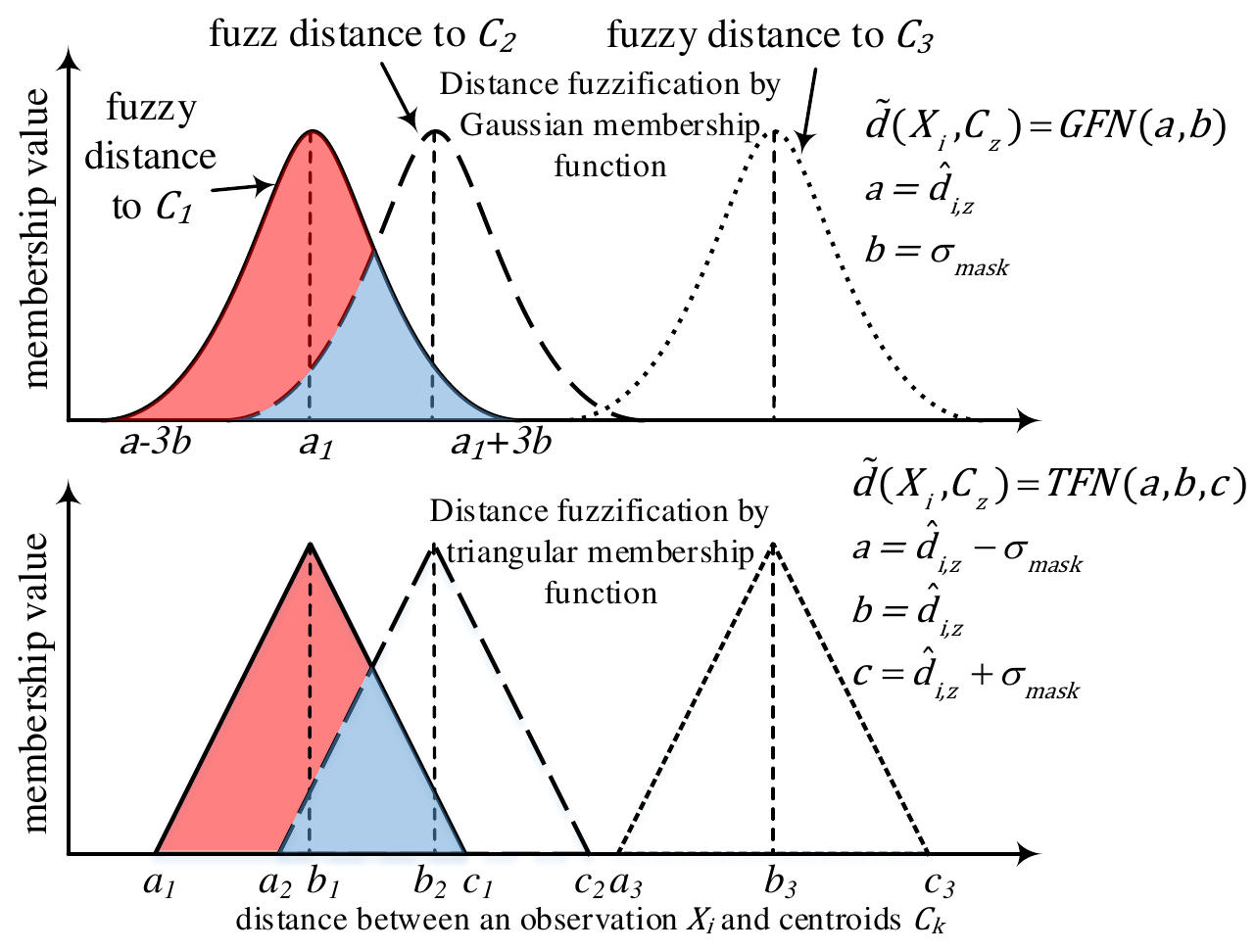}
    \caption{
    A demonstration of Gaussian fuzzy distance and triangular fuzzy distance. For an observation $X_i$ with missing values, the distances between $X_i$ and the centroids of the clusters $C_1$, $C_2$, $C_3$ are fuzzified. In this case, $X_i$ might belong to both the $C_1$ and $C_2$ clusters but not to $C_3$ since the upper bound of the estimated distance $\tilde{d}(X_i,C_1 )$ is less than the lower bound of $\tilde{d}(X_i,C_1 )$ that is $\overline{d} (X_i,C_1 ) < \underline{d} (X_i,C_3 )$. For simplicity, we use $\tilde{d}_{i,1}$ to denote $\tilde{d}(X_i,C_1 )$, where the fuzzy distance can be presented by Gaussian fuzzy number ($\mathrm{GFN}(a, b)$) and Triangular fuzzy number ($\mathrm{TFN}(a, b, c)$).
    }
    \label{fig:5}
\end{figure}

After building the missing value distance prediction model $L_{mask}$ and obtaining the standard deviation $\sigma_{mask}$ of the errors, we can now start estimating the frequency of observations in histogram bins.
For any two observations with missing values, we use the same masking function to convert them then pass them to $L_{mask}$. The returned result $\hat{d}$ with the standard deviation $\sigma_{mask}$ will be used to construct the fuzzy distance to present its membership to a bin. In fact, there are two options with the predicted distance $\hat{d}$. One is simply to use the predictions $\hat{d}$ as crisp distances without considering the errors of $L_{mask}$. The observations would then be assigned to histogram bins in a many-to-one manner, which means each observation would only be counted once by a bin. Another option is to fuzzify the distance according to the $\sigma_{mask}$, that is, to consider the estimated distance as a fuzzy number, as shown in Fig. \ref{fig:5}. 

The workflow of fuzzy-weighted frequency is as follows. Step 1) model the estimated distance between $X_i$ and $C_z$ as fuzzy distance, $\tilde{d}_{i,z}$;  Step 2) calculate the degree of membership $\mu_{i,k}$ of every observation to each bin based on the fuzzy distances. Step 3) normalize the degree of membership as observation's weight. Step 4) Accumulate the weights in each bin as the frequency. The novelty of the FWF compare to fuzzy clustering is that the degree of membership is calculated based on fuzzy distance rather than crisp distance.

In this case, the sample set $X$ is the universe of discourse. The value $\mu(X_i )$ for each $X_i\in X$ is the grade of membership of $X_i$ in $(X,\mu)$. The function $\mu(X_i )$ is the membership function of a fuzzy set $C=(X,\mu)$, where a fuzzy set is a cluster denoted as $C_k$.  When measuring the frequency of observations, the fuzzy distance will be used to determine the degree of membership of an observation to each clusters.

We propose two types of fuzzy numbers to present the estimated distance: the Triangular fuzzy numbers (TFN) and Gaussian fuzzy number (GFN). We chose the TFN as a baseline for comparison because it is the most widely used \cite{yiliao:TFS}, and we chose the GFN as it is the most reasonable since the predictions and errors are assumed to follow a Gaussian distribution. Once the fuzzy distance formulation is determined, we can leverage them to calculate the degree of membership. The contingency table is then built with the degree of membership.
\begin{figure*}[ht]
    \centering
    \includegraphics[scale=0.8]{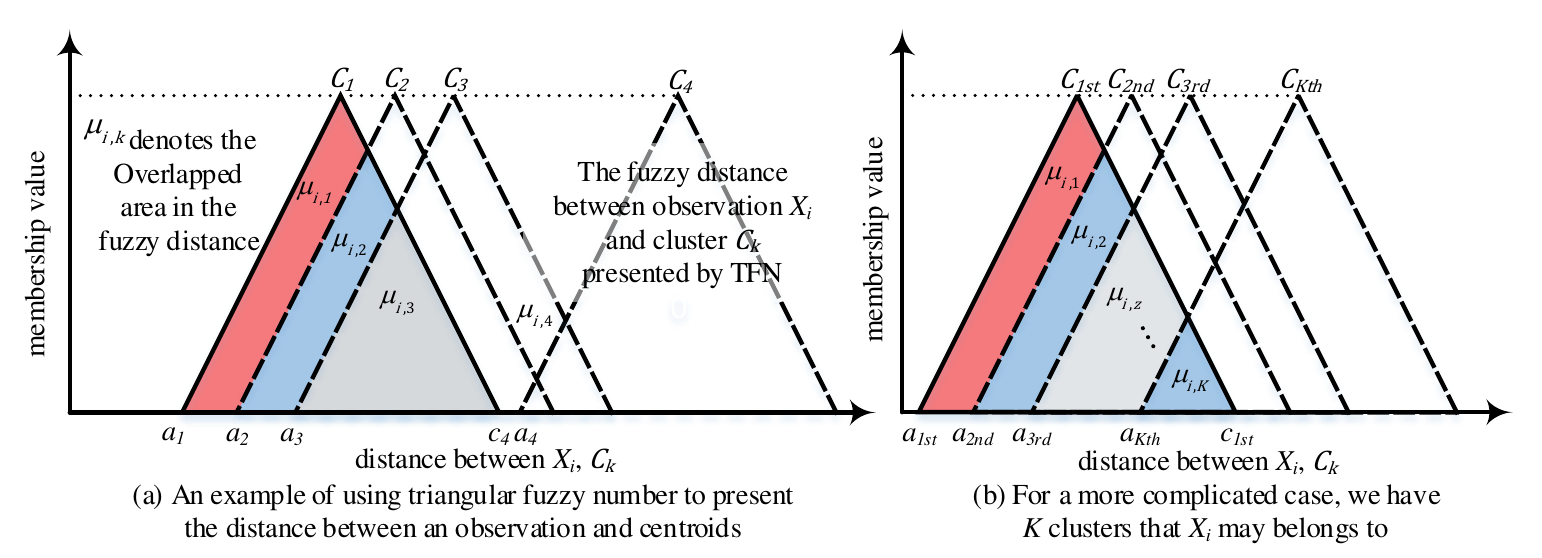}
    \caption{
    An illustration of normalizing the degree of membership. $K$ denotes the total number of clusters. $\mu_{i,z}$ denotes the overlapping triangular area used for the weight calculation.
    }
    \label{fig:6}
\end{figure*}

In general, the membership degree of an observation with missing values to a cluster is determined by its estimated distances to the cluster centroids and the standard deviation of the errors. 
This is formulated by the membership function $\mu\left(X_i, C_z\right)$, namely $\mu_{i,z}$. Without any prior knowledge of the sample set, we assume the observations all have equal weights of $1$. Therefore, the fuzzy weighting function can be formulated as a normalized degree of membership, denoted as
\begin{equation}
\label{equ:fwf}
    w_{C_z}(X_i)=\frac{\mu_{i,z}}{\sum_{k=1}^K \mu_{i,k}}.
\end{equation}
Fig. \ref{fig:6} illustrates how to calculate the weights based on the Triangular fuzzy distance. In this example, the estimated distance between an observation $X_i$ and the cluster centroids $\{C_k\}_{k=1}^4$ are represented with a Triangular fuzzy number.  As shown, $X_i$ is closest to the centroid of $C_1$ and furthest from the centroid of $C_4$. Even considering a possible error by $L_{mask}$, it is still very unlikely that $X_i$ should fall into cluster $C_4$, which can be formulated by the condition that $a_4\geq c_1$. 
To calculate how likely it is that observation should fall into $C_1$, $C_2$, $C_3$ and $C_4$, we use the overlapping areas as a membership function. The most likely cluster for $X_i$ is $C_1$ with the overlapping area $\mu_{i,1}$, as the areas for $\mu_{i,2}$ and $\mu_{i,3}$ are smaller, and $\mu_{i,4}=0$. The sum of the total overlapped area is $\mu_{i,1}+\mu_{i,2}+\mu_{i,3}+\mu_{i,4}$. Therefore, we can normalize the weight of $X_i$ to each cluster as $w_{C_1} (X_i )=\frac{\mu_{i,1}}{\mu_{i,1}+\mu_{i,2}+\mu_{i,3}+\mu_{i,4}}$, $w_{C_2} (X_i )=\frac{\mu_{i,2}}{\mu_{i,1}+\mu_{i,2}+\mu_{i,3}+\mu_{i,4}}$, $w_{C_3} (X_i )=\frac{\mu_{i,3}}{\mu_{i,1}+\mu_{i,2}+\mu_{i,3}+\mu_{i,4}}$, and $w_{C_4} (X_i )=0$.


The degree of membership of $\mu_{i,k}$ is calculated based on the fuzzy distance. We define the triangular fuzzy distance as a $\mathrm{TFN}(a,b,c)$, that is, 
\begin{equation*}
    \begin{cases}
      a=\hat{d}_{i,z}-\sigma_{mask} \\
      b=\hat{d}_{i,z} \\
      c=\hat{d}_{i,z}+\sigma_{mask} \\
    \end{cases},
\end{equation*}
where $\hat{d}_{i,z}$ is the predicted distance between observation $X_i$ and the cluster $C_z$.
\hlc{
Since all triangles are congruent triangles, the triangles of $\mu_{i,z}$ and $\mu_{i,1}$ must be similar. Hence, for simplicity in explaining how the degrees of membership are normalized, we have assumed that the degree of membership of an observation to the closet cluster is equal to 1, i.e., $\mu_{i,1}=1$. Based on the area theorem of similar triangles, if two triangles are similar, then the ratio of the area of both triangles is proportional to the square of the ratio of their corresponding sides. Therefore, the degree of membership of observation $i$ to its $z$-th closest cluster is}
\begin{equation}
\label{equ:tri_mu}
    \begin{aligned}
    \mu_{i,z}& =\left(\frac{c_1-a_z}{c_1-a_1}\right)^2\mu_{i,1} \\
    & = \left(\frac{\hat{d}_{i,1} + \sigma_{mask} - (\hat{d}_{i,z}-\sigma_{mask} )}{\hat{d}_{i,1} + \sigma_{mask} - (\hat{d}_{i,1}-\sigma_{mask})}\right)^2 \\
    & =\frac{(2\sigma_{mask}+\hat{d}_{i,1}-\hat{d}_{i,z})^2}{4\sigma_{mask}^2}
    \end{aligned},
\end{equation}
\hlc{where $\left(\frac{c_1-a_z}{c_1-a_1}\right)^2$ is the square of the ratio of their corresponding sides.}
Substitute to Eq. \eqref{equ:fwf} we have the triangular fuzzy-weighted frequency represented as:
\begin{equation}
\label{equ:tri_detail}
    w_{C_z}^{Tri}(X_i)=\frac{(c_1-a_z)^2}{Kc_1^2-2c_1\sum_{i=1}^K  a_i + \sum_{i=1}^K a_i^2}.
\end{equation}
Because of their smoothness and concise notation, Gaussian and bell membership functions are popular methods for specifying fuzzy sets. With these functions, the distance between the cluster centroids and the observations is a Gaussian fuzzy number, $\mathrm{GFN}(a, b)$, and the prediction value $\hat{d}$ plus standard deviation $\sigma_{mask}$ are used to model the distance, that is,
\begin{equation*}
    \begin{cases}
      a=\hat{d}_{i,z}\\
      b=\sigma_{mask} \\
    \end{cases}.
\end{equation*}
The degree of membership of $X_i$ to $C_z$ is the area under the curve and is formulated as
\begin{equation}
\label{equ:gau_mu}
\begin{aligned}
    \mu_{i,z}&= 2 \left(\frac{1}{2}\left[ 1+\mathrm{erf}\left( \frac{\frac{a_z+a_1}{2}-a_z}{\sqrt{2}b_z}\right)\right]\right)\\
    &=1+\mathrm{erf}\left( \frac{\frac{\hat{d}_{i,z}+\hat{d}_{i,1}}{2}-\hat{d}_{i,z}}{\sqrt{2}\sigma_{mask}}\right)\\
    &=1+\mathrm{erf}\left( \frac{\hat{d}_{i,1}-\hat{d}_{i,z}}{2\sqrt{2}\sigma_{mask}}\right)
     \end{aligned}.
\end{equation}
And, similar to the triangular fuzzy number, the Gaussian fuzzy-weighted frequency is
\begin{equation}
\label{equ:gau_detail}
    w_{C_z}^{Gau}(X)=\frac{1+\mathrm{erf}\left( \frac{a_1-a_z}{2\sqrt{2}b_z}\right)}{K+\sum_{i=1}^K\mathrm{erf}\left( \frac{a_1-a_i}{2\sqrt{2}b_i}\right)}.
\end{equation}
In contrast to distance fuzzification, the crisp frequency is measured based on $\hat{d}$, and the frequency measuring is proceeded by finding the closest cluster.

To simplify the computation and to avoid the standard deviation $\sigma_{mask}$ become too large to provide reasonable fuzzy distances. We propose a Top-Q cluster constrain for normalization the weight, that is, only the closest $Q$ clusters will be considered for frequency measuring, that is, the Eq. \eqref{equ:fwf} is updated as follow.
\begin{equation}
\label{equ:topq}
    w_{C_z}(X_i, Q) = \frac{\mu_{i,z}}{\sum_{k=1}^Q \mu_{i,k}}I_{z\leq Q}.
\end{equation}
where, $Q\leq K$, the $I_{z\leq Q}$ is the indicator function that $I_{z\leq Q}=1$ if $z\leq Q$, otherwise $0$. When $Q=1$, the measured frequency will be equal to the crisp distance frequency. The Alg \ref{alg:2} presents the pseudocode of fuzzy-weighted frequency measuring. With the frequency, we can apply the chi-square test to accept or reject the null hypothesis.

\hlr{
Alg. \ref{alg:2} presents the pseudocode of the fuzzy-weighted frequency. The runtime complexity is determined by $k$-means clustering (Line 1-3), imputation (Line 4-8) and measuring the fuzzy-weighted frequency (Line 9-15). For the clustering, according to Lloyd's $k$-means algorithm, the runtime complexity of the clustering procedure in our algorithm is $\mathcal{O}(m^2n)$.  The complexity of imputing the data is based on the worst case of the imputation algorithm in $\Xi$. Given our default settings, the worst case is an iterative imputation with $\mathcal{O}(m^2n)$ on a $m\times n$ data set. The fuzzy weighting process has linear runtime complexity of $\mathcal{O}(mn)$. Therefore, the runtime complexity of Alg. \ref{alg:2} is $\mathcal{O}(m^2n)$. Combining Algs. \ref{alg:1} and \ref{alg:2}, the overall runtime complexity of MDL-FWF is $\mathcal{O}(n_{split}m^2n + m^2n)=\mathcal{O}(n_{split}m^2n)$, which is bounded by the MDL algorithm.}

\begin{algorithm}[ht]
    \caption{Fuzzy-weighted Frequency (FWF)}
    \label{alg:2}
    \small
    \SetKwInOut{Input}{input}
    \SetKwInOut{Output}{output}

    \Input{1. Observations, $X$
        \newline 2. Data imputation algorithm pool, default $\Xi=(\xi_{zero},\xi_{mean},\xi_{med},\xi_{mfr},\xi_{iter})$
        \newline 3. Number of clusters (bins), default $K=m/50$
        \newline 4. Top-Q constrain, default $Q=3$
        \newline 5. Trained distance learning model, $L_{mask}$
        \newline 6. Prediction error, $\sigma_{mask}$
        \newline 7. Distance fuzzification model, default TFN
        }
    \Output{1. Frequency vector, $V_k$
    \newline 2. Cluster centroids, $\mathcal{C}=\{C_k\}_{k=1}^K$
    }
    \BlankLine
    retrieve non-missing observations, $X^{complete}$\;
    $k$-means cluster $X^{complete}$ and get the centroids $\{C_k\}_{k=1}^K$\;
    retrieve missing observations, $X^{miss}$\;
    \For {$\xi_i$ $\mathrm{in}$ $\Xi$} {
    	apply data imputation $X^{impu}=\xi(X^{miss})$\;
    	calculate distances between $X^{impu}$ and $C_k$ as $x_{2n+i}$\;
    	concatenate horizontally $x_{2n+i}$ to $X^{miss}$\;
    }
    predict distance on $X^{miss}$, $\hat{d} = L_{mask}(X^{miss})$\;
   initial  frequency vector $V_k=\{0\}_{1,\ldots,K}$\;
    \For {$v_k$ in $V_k$ } {
      \For {$X_i$ in $X$ } {
          update frequency $v_k\leftarrow v_k+w_{C_k}(X_i)$ \tcp*[r]{Eqs. \eqref{equ:topq} \eqref{equ:tri_mu}\eqref{equ:tri_detail} or \eqref{equ:gau_mu}\eqref{equ:gau_detail} depends on the membership function}
      }
    }
    \Return $V_k$, $\mathcal{C}$
\end{algorithm}


\section{Experiments and Evaluation}
\label{s:IV}

\hly{In this section, we evaluate the proposed algorithms in three respects. Section \ref{ss:exp1} evaluates the distance estimation algorithm in comparison to various data imputation methods. Section \ref{ss:exp2} tests MDL-FWF's performance at concept drift detection on five synthetic data sets with MCAR missing values. In this set of experiments, we also examine whether fuzzifying the observations improves performance. Section \ref{ss:exp3} and \ref{ss:exp4} present comparisons with five state-of-the-art concept drift detection algorithms.}

\subsection{Masked Distance Learning vs. Calculating Distance from Imputed Values}
\label{ss:exp1}
\noindent\textbf{Experiment 1 - Distance estimation with MCAR}.

This experiment evaluate the extent to which masked distance learning can reduce estimation errors.

\noindent\textbf{Data sets}.

\begin{table}[ht]
\centering
\caption{Synthetic data set characteristics for distance estimation.}
\label{tab:I}
\begin{tabular}{lll}
\toprule
Distribution & Distribution Config                                   & MV Config                       \\
\midrule
Uniform      & $x_i\in[0,10],X=\{x_i\}_{i=1}^{10}$                       & \begin{tabular}[c]{@{}l@{}}$V_{r_{1,\ldots,5}}=0.2$\\ $V_{r_{6,\ldots,10}}=0.0$\end{tabular} \\
Gaussian     & $\mu=\mathrm{rand}(0,1)$, $\sigma=\mathrm{rand}(0,1)$ & Same as above                                                              \\

Exponential  & $\beta=\mathrm{rand}(0,1)$                     & Same as above                                                              \\
Poisson      & $\lambda=\mathrm{randint}(5,10)$                             & \begin{tabular}[c]{@{}l@{}}$V_{r_{1,\ldots,3}}=0.2$\\ $V_{r_{4,5}}=0.0$\end{tabular}  \\
Categorical  & $\mathrm{randint}(0,10)$                                        & Same as above                             \\
\bottomrule
\end{tabular}
\end{table}
For each distribution, we generated 500 observations and randomly removed some values in different configurations, as summarized in Table \ref{tab:I}. For the uniform, Gaussian, and exponential distributions, we generated $n=10$ dimensional data. The first $5$ dimensions had a missing value rate of 20\%. For the Poisson, and uniform categorical distributions, we generated $n=5$ dimensional data where the first $3$ dimensions had a missing value rate of 20\%. 

\noindent\textbf{Imputation methods and configurations}.

Missing values were replaced with five kinds of imputed values: zeros, means, medians, most frequent, and iterative. We generated five sample sets with different distributions and stored the true pairwise distance matrix for each. Then we randomly masked some values as null and applied different distance estimation methods to estimate the pairwise distance matrix. The error between the estimated distance and the true distance was used to evaluate performance.

\noindent\textbf{Evaluation metric}.

We used both root mean squared error (RMSE) and mean absolute error (MAE) as the metric for distance estimation evaluation.
RMSE gives a relatively high weight to large errors. This means the RMSE should be more useful when large errors are undesirable. 
MSE does not penalize huge errors as badly as RMSE, but it is a good metric to reflect the overall prediction accuracy.

\begin{table*}[ht]
\centering
\caption{\hlc{Results of the missing value estimation. The MAE and RMSE are calculated based on the difference between the true values and the estimated values.}}
\label{tab:value_diff}
\begin{tabular}{llllllllll}
\toprule
    &      & impute\_zero     & impute\_mean             & impute\_medi             & impute\_mfre    & impute\_iter             & MIDAS            & GAIN            & MDL \\ \midrule
Uni & MAE  & 0.071$\pm$0.012  & \textbf{0.036$\pm$0.006} & \textbf{0.036$\pm$0.006} & 0.071$\pm$0.012 & \textbf{0.036$\pm$0.006} & 0.042$\pm$0.008  & 0.039$\pm$0.007 & NA  \\
    & RMSE & 0.217$\pm$0.022  & 0.109$\pm$0.011          & \textbf{0.012$\pm$0.041} & 0.216$\pm$0.022 & 0.109$\pm$0.011          & 0.133$\pm$0.016  & 0.125$\pm$0.016 & NA  \\
Gau & MAE  & 0.802$\pm$0.116  & \textbf{0.113$\pm$0.020} & \textbf{0.113$\pm$0.020} & 0.418$\pm$0.065 & \textbf{0.113$\pm$0.020} & 0.442$\pm$0.077  & 0.133$\pm$0.029 & NA  \\
    & RMSE & 2.189$\pm$0.176  & 0.372$\pm$0.048          & \textbf{0.141$\pm$0.636} & 1.177$\pm$0.113 & 0.373$\pm$0.048          & 1.294$\pm$0.142  & 0.440$\pm$0.073 & NA  \\
Exp & MAE  & 0.105$\pm$0.021  & 0.105$\pm$0.021          & \textbf{0.100$\pm$0.023} & 0.143$\pm$0.030 & 0.105$\pm$0.021          & 0.104$\pm$0.021  & 0.112$\pm$0.023 & NA  \\
    & RMSE & 0.370$\pm$0.069  & 0.371$\pm$0.069          & \textbf{0.156$\pm$1.243} & 0.527$\pm$0.088 & 0.371$\pm$0.069          & 0.370$\pm$0.069  & 0.410$\pm$0.070 & NA  \\
Poi & MAE  & 6.417$\pm$1.145  & 0.975$\pm$0.200          & \textbf{0.974$\pm$0.200} & 1.025$\pm$0.216 & 0.976$\pm$0.199          & 5.629$\pm$1.101  & 1.179$\pm$0.278 & NA  \\
    & RMSE & 16.002$\pm$1.749 & \textbf{2.524$\pm$0.404} & 6.566$\pm$27.245         & 2.672$\pm$0.450 & 2.529$\pm$0.404          & 14.989$\pm$1.710 & 3.081$\pm$0.667 & NA  \\
Cat & MAE  & 0.643$\pm$0.134  & 0.293$\pm$0.053          & \textbf{0.285$\pm$0.056} & 0.342$\pm$0.075 & 0.293$\pm$0.053          & 0.579$\pm$0.123  & 0.320$\pm$0.068 & NA  \\
    & RMSE & 1.713$\pm$0.248  & 0.694$\pm$0.092          & \textbf{0.544$\pm$1.884} & 0.925$\pm$0.149 & 0.695$\pm$0.092          & 1.566$\pm$0.242  & 0.823$\pm$0.140 & NA \\\bottomrule
\end{tabular}
\end{table*}

\begin{table*}[ht]
\centering
\caption{\hlc{Results of the distance estimation of two observations with missing values. The compared methods are imputation by zero, mean, median, most frequent, iterative imputation, MIDAS, GAIN and masked distance learning. The best results appear in bold.}}
\label{tab:II}
\begin{tabular}{llllllllll}
\toprule
            &      & impute\_zero     & impute\_mean    & impute\_medi                              & impute\_mfre    & impute\_iter     & MIDAS            & GAIN            & MDL                                       \\ \midrule
Uni     & MAE  & 0.146$\pm$0.003  & 0.087$\pm$0.003 & 0.087$\pm$0.003                           & 0.145$\pm$0.003 & 0.087$\pm$0.003  & 0.087$\pm$0.003  & 0.087$\pm$0.003 & \textbf{0.077$\pm$0.002} \\
            & RMSE & 0.202$\pm$0.003  & 0.127$\pm$0.004 & 0.127$\pm$0.004                           & 0.201$\pm$0.004 & 0.127$\pm$0.004  & 0.127$\pm$0.003  & 0.127$\pm$0.004 & \textbf{0.104$\pm$0.002} \\
Gau    & MAE  & 3.731$\pm$0.031  & 0.294$\pm$0.015 & 0.294$\pm$0.015                           & 1.247$\pm$0.122 & 0.3294$\pm$0.015 & 1.456$\pm$0.132  & 0.304$\pm$0.026 & \textbf{0.280$\pm$0.014} \\
            & RMSE & 4.429$\pm$0.034 & 0.469$\pm$0.024 & 0.469$\pm$0.024                           & 1.598$\pm$0.147 & 0.469$\pm$0.024  & 1.959$\pm$0.145  & 0.473$\pm$0.034 & \textbf{0.403$\pm$0.019} \\
Exp & MAE  & 0.295$\pm$0.025  & 0.296$\pm$0.025 & \textbf{0.289$\pm$0.026} & 0.314$\pm$0.023 & 0.296$\pm$0.025  & 0.295$\pm$0.025  & 0.297$\pm$0.025 & 0.309$\pm$0.026                           \\
            & RMSE & 0.572$\pm$0.069  & 0.572$\pm$0.069 & 0.575$\pm$0.069                           & 0.583$\pm$0.060 & 0.572$\pm$0.069  & 0.572$\pm$0.069  & 0.570$\pm$0.067 & \textbf{0.542$\pm$0.072} \\
Poi     & MAE  & 21.252$\pm$0.204 & 2.214$\pm$0.112 & 2.214$\pm$0.112                           & 2.242$\pm$0.113 & 2.215$\pm$0.112  & 19.098$\pm$0.241 & 2.396$\pm$0.300 & \textbf{2.073$\pm$0.098} \\
            & RMSE & 29.043$\pm$0.230 & 3.803$\pm$0.203 & 3.805$\pm$0.202                           & 3.834$\pm$0.200 & 3.803$\pm$0.203  & 26.766$\pm$0.302 & 4.018$\pm$0.451 & \textbf{3.284$\pm$0.164} \\
Cat & MAE  & 1.390$\pm$0.029  & 0.648$\pm$0.029 & 0.638$\pm$0.029                           & 0.692$\pm$0.057 & 0.648$\pm$0.029  & 1.221$\pm$0.026  & 0.660$\pm$0.041 & \textbf{0.598$\pm$0.027} \\
            & RMSE & 2.168$\pm$0.042  & 1.046$\pm$0.051 & 1.048$\pm$0.050                           & 1.131$\pm$0.097 & 1.046$\pm$0.051  & 1.945$\pm$0.042  & 1.069$\pm$0.067 & \textbf{0.915$\pm$0.049} \\ \bottomrule
\end{tabular}
\end{table*}

\begin{figure}[ht]
    \centering
    \includegraphics[scale=0.4]{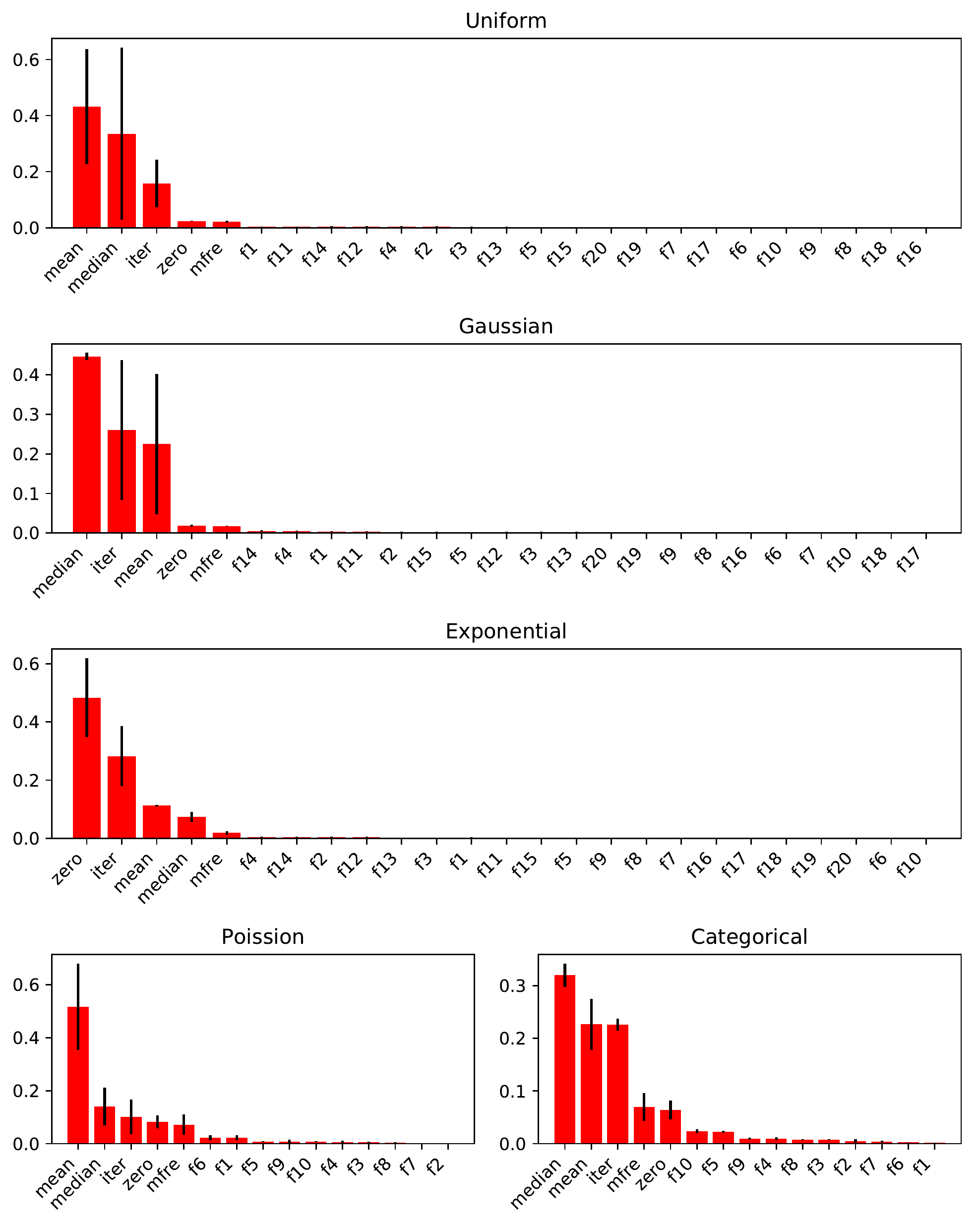}
    \caption{
    Feature importance returned by gradient boosting decision tree, calculated based on toatl gain. For different data sets. The $f_i$s are the original features. For Uniform distribution, an observation $X_1$ has 10 features. To compute the paired distance, for example, the distance between $X_1$, $X_2$, we horizontally concatenated $X_1$ and $X_2$, and append the distance calculated based on imputations. Therefore, the input feature space for $L_{mask}$ are $f_1,\ldots, f_{20}$ and $mean,~median,~zero,~mfre,~iter$. For Poission and Categorical data set, there were 5 features for the original observations, therefore, the concatenated features are $f_1,\ldots, f_{10}$ and $mean,~median,~zero,~mfre,~iter$
    }
    \label{fig:7}
\end{figure}
\noindent\textbf{Findings and discussion}.

The distance estimation results are shown in Table \ref{tab:II}, and the feature importance is plotted in Fig. \ref{fig:7}. In this experiment, we find that combining multiple imputation methods and training a regression model improves the overall estimations. It is easy to understand that regression considers the different imputation method will reach the best result. Compared to calculating distance from imputed values, using masked distance learning to predict the distance had lower RMSE and MAE. This inspired us to think that combining multiple instances of uncertainty into one problem and searching for the best optimization solution may produce a better result than solving each uncertainty issue individually. 

Regarding the feature importance returned by the gradient boosting decision tree, we noticed that different imputation methods resulted in different importance ranks for different distributions. For example, mean imputation contributed the most for the distance etismtaion with missing values on uniform and Poisson distributions while, for exponential distributions, zero imputation was more important. The overall estimation results show that integrating diverse imputation method for distance estimation with missing values is effectual.

\subsection{Fuzzy-weighted Frequency vs. Crisp Frequency Drift Detection}
\label{ss:exp2}
\begin{figure*}[ht]
    \centering 
    
    \subfloat[Uniform Mean Drift]{\includegraphics[width=0.32\textwidth]{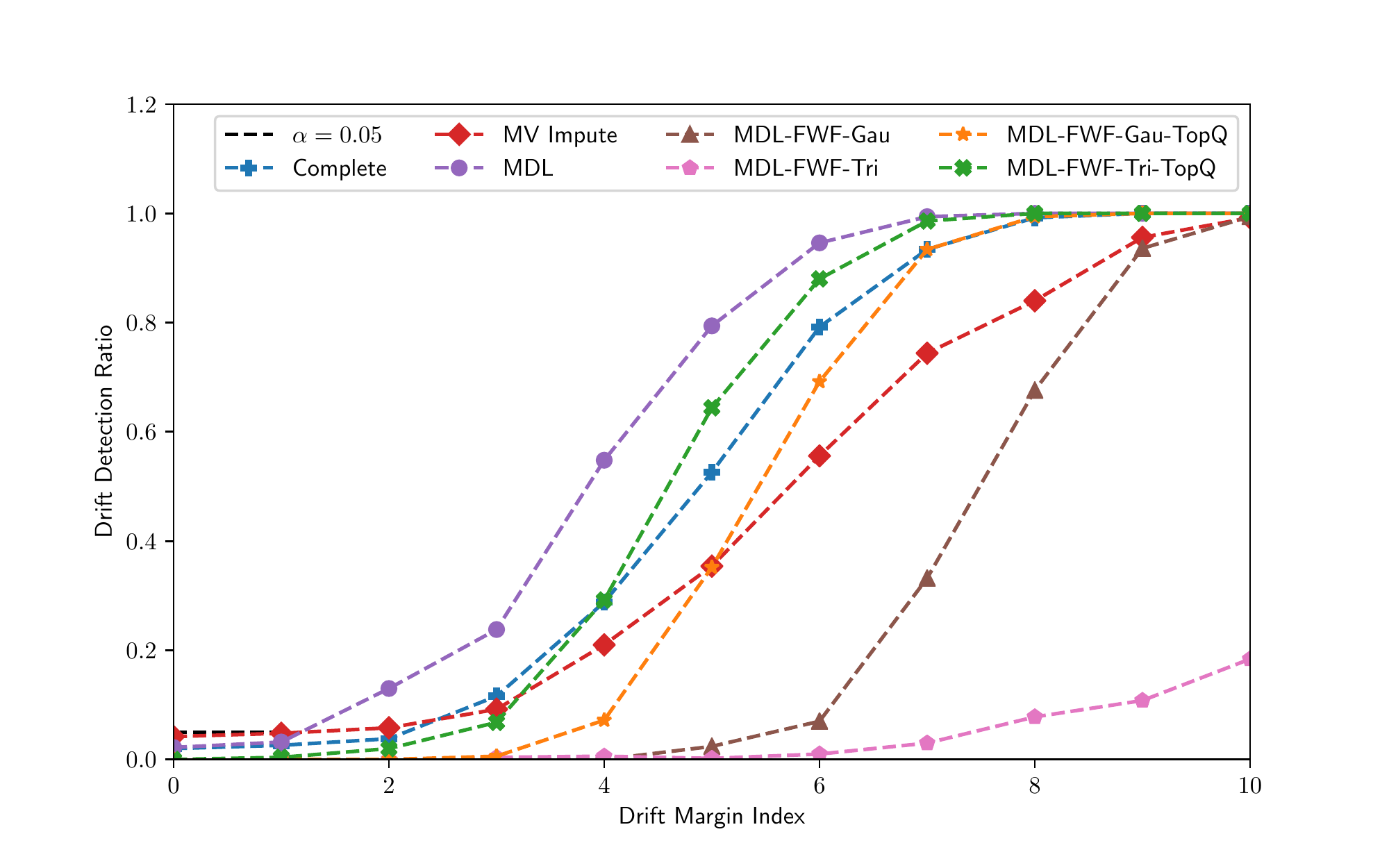}}
    \subfloat[Gaussian Mean Drift]{\includegraphics[width=0.32\textwidth]{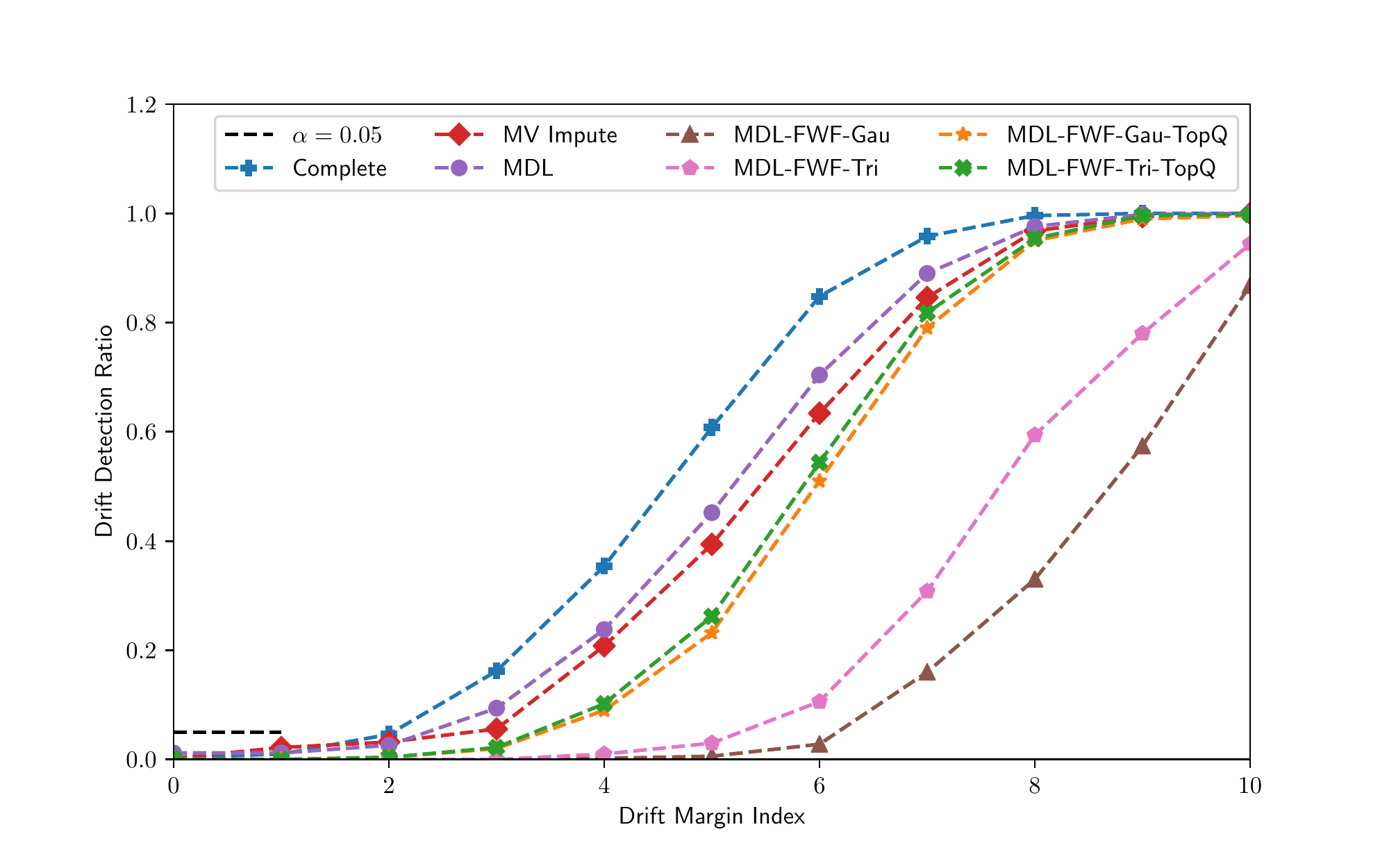}}
    \subfloat[Gaussian Covariance Drift]{\includegraphics[width=0.32\textwidth]{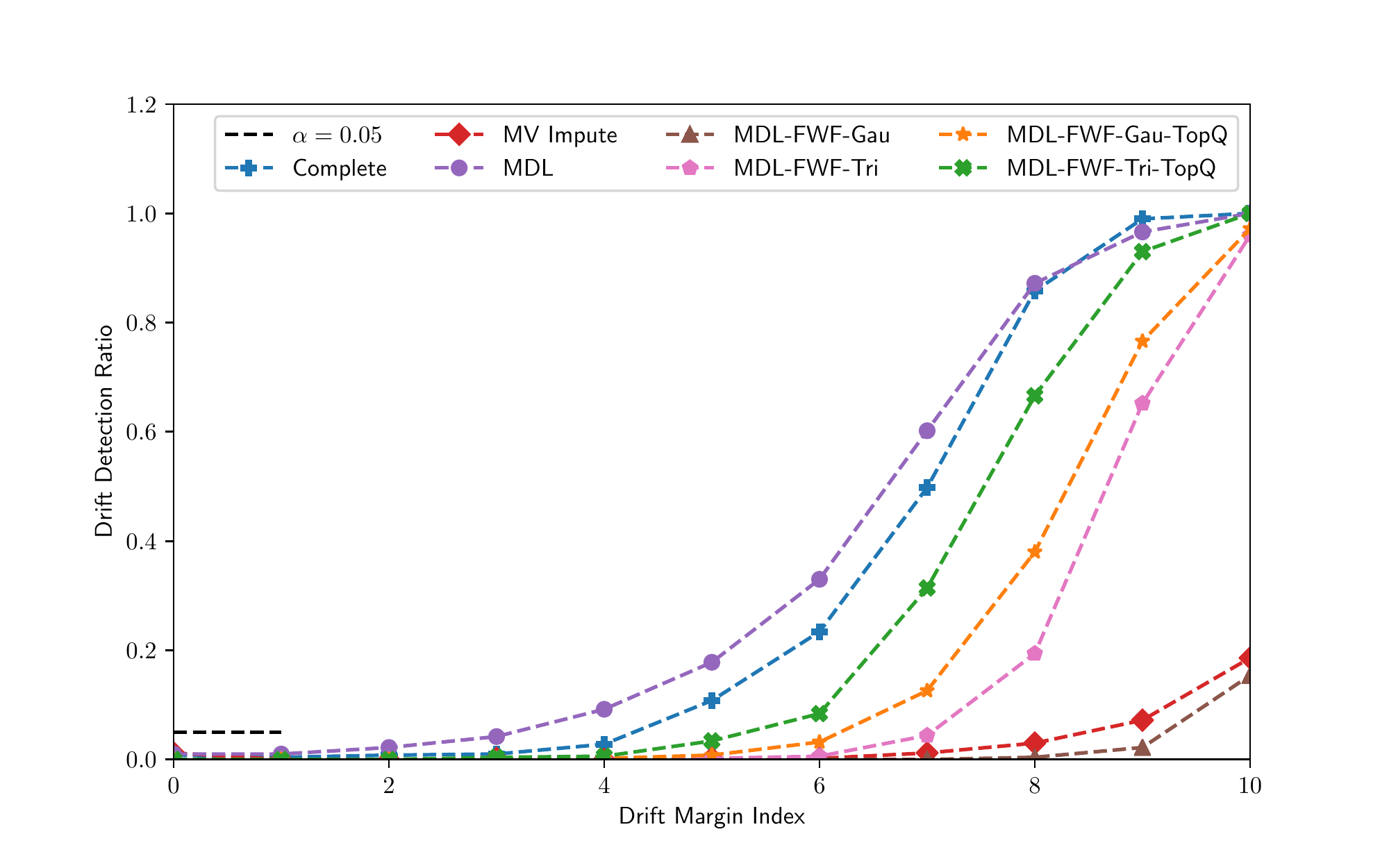}}
    \\
    \subfloat[Poisson Mean Drift]{\includegraphics[width=0.32\textwidth]{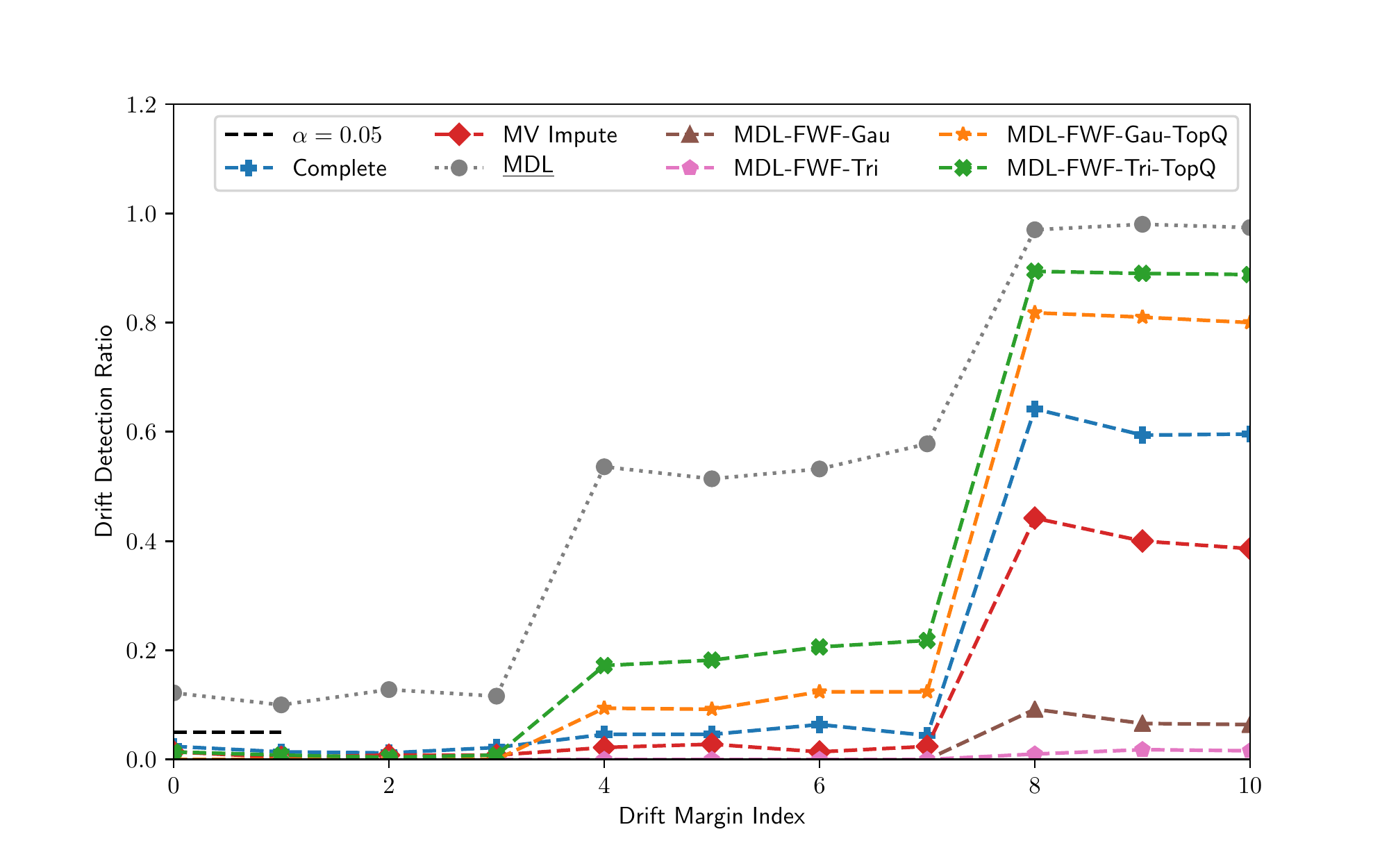}}
    \subfloat[Poisson $\rho$ Drift]{\includegraphics[width=0.32\textwidth]{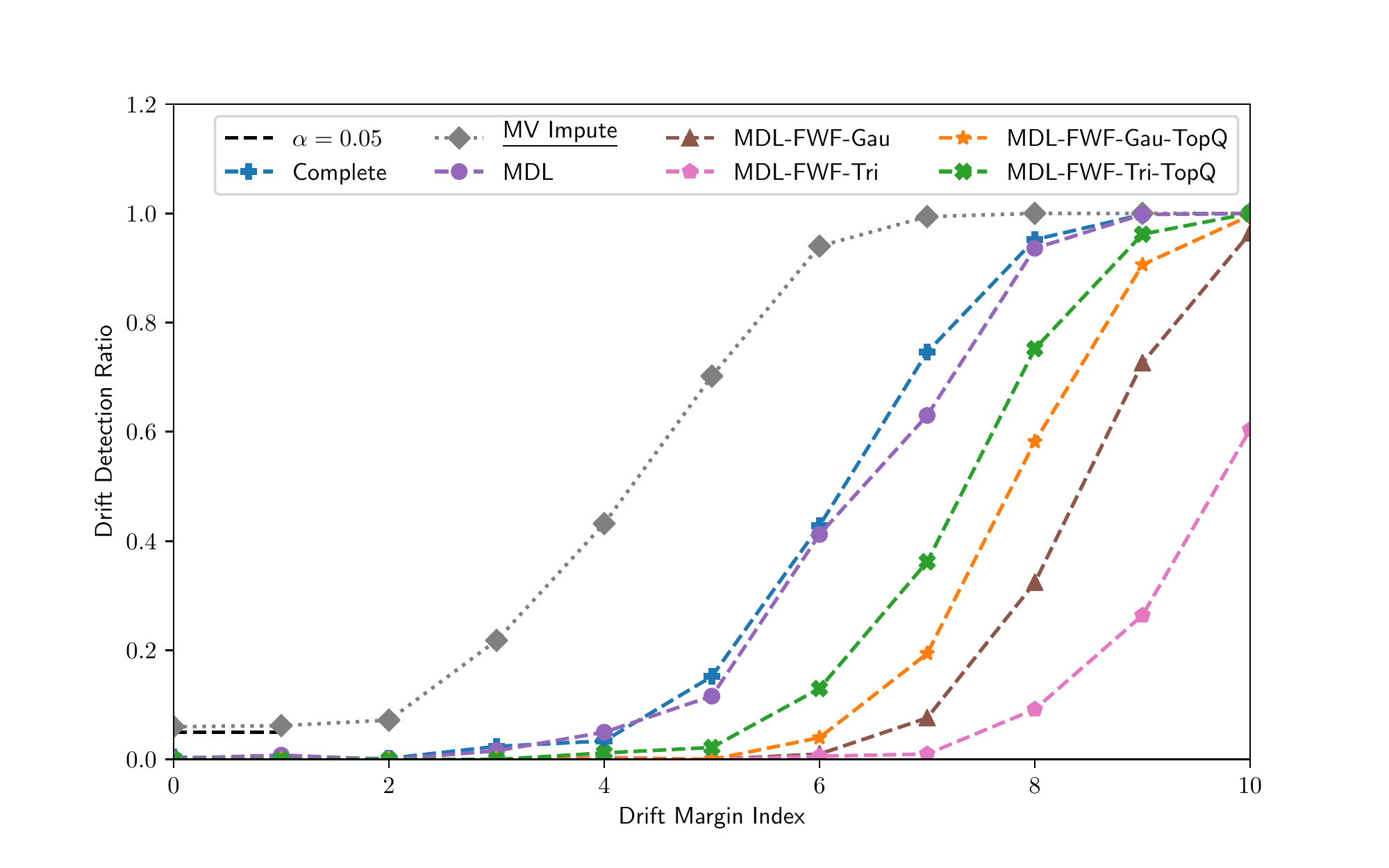}}
    \subfloat[Average Result of Ablation Test]{\includegraphics[width=0.32\textwidth]{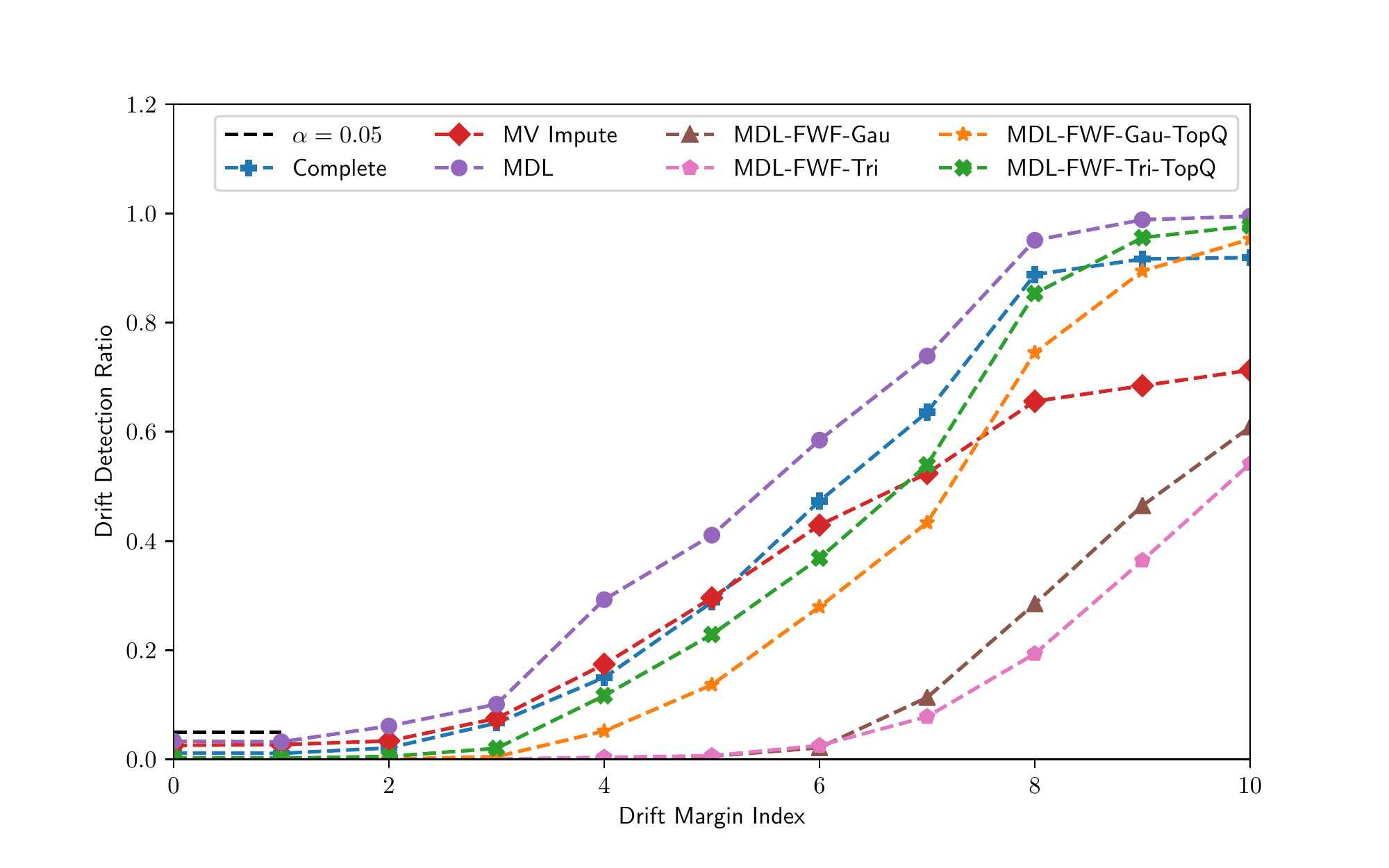}}
    
    \caption{The drift detection ratios for different components of MDL-FWF on synthetic data sets with different drift margins. The average drift detection ratio was calculated based on the drift margin index. The margin index indicate the severity of the drift. The higher the index, the more severe the drift. We used the average to demonstrate how performance changed with different severities of drift.}
    \label{fig:8}
\end{figure*}

\noindent\textbf{Experiment 2 – Algorithm ablation test}

In this experiment, we analyzed the contribution of each component's performance of MDL-FWF.

\noindent\textbf{Data sets}.

First, we needed to simulate drift, so we generated data sets using three different distributions and five different types of drift. \hlr{We adopt the commonly used drift synthetic methods for drift detection \cite{Lu:AI1, Dasu:kdqTree, Boracchi:QTree, MWWtest}}. 

\noindent\textbf{Configurations}.

Config. 1 – No missing values (Complete). The first configuration is the MDL-FWF baseline, i.e., drift detection with no missing values in the data set. In this configuration, we created k-Means space partitions and counted the observations in each cluster as the frequency. The presence of a drift was determined from a Pearson’s chi-square test. 

Config. 2 – Impute missing values (MV Impute). In this setting, we duplicated the data set generated in Config. 1 and introduced MCAR missing values by randomly replacing some values with null. Those missing values were then imputed with iterative imputation. We chose iterative imputation because after comparing the five imputation methods, the iterative approach gave the best performance except MDL. k-Means and Pearson’s chi-square tests were used to detect drifts.

Config. 3 – Masked distance learning only (MDL). In this configuration, we used the masked distance learning algorithm to estimate the distance between the cluster centroids and the observations with missing values. Those estimated distances were then used to compute the contingency table. Again, Pearson’s chi-square test was used to detect drifts.

Configs. 4.1 \& 4.2 – MDL-FWF. Similar to Config. 3, we predicted the distance between the cluster centroids and the observations with missing values via MDL. However, as opposed to using the estimated distances to build the contingency table, we used the fuzzy-weighted frequency measurements via a Gaussian fuzzy membership function (Config. 4.1) and a triangular fuzzy membership function (Config. 4.2). Pearson’s chi-square tests determined drift.

Configs. 5.1 \& 5.2 – MDL-FWF with the top-Q clusters (MDL-FWF-TopQ). Based on Configs. 4.1 and 4.2, we measured the fuzzy-weighted frequency of the top-Q clusters, instead of considering the degree of the membership to all clusters. The same membership functions as in Configs. 4.1 and 4.2 apply to Configs. 5.1 and Config. 5.2.

\noindent\textbf{Evaluation metric}.

The conventional drift detection evaluation metrics are Type-I and Type-II errors. Type-I errors are rejections of a true null hypothesis (also known as “false positives”), and Type II errors are the false null hypothesis rates (i.e., “false negatives”). For concept drift detection, the null hypothesis is that, there is no concept drift between the populations. Commonly, the Type-I and Type-II errors are shown separately and it is difficult to visualise them in one plot. Therefore, to better visualize the results of each drift detection method, we used the drift detection ratio as the metric \cite{Liu:PR}, which is the number of rejected hypotheses divided by the total number of tests.

To quantify the performance of each configuration, we used Pearson’s correlation coefficient as an indicator of the similarity between different drift detection configurations. Since the base model is kMeans, the drift detection ratios of Config. 1 (No MV) forms the baseline and Configs. 2 to 5 show the impact of each component. A higher Pearson’s correlation coefficient to Config. 1 means that the component is less affected on the missing values. 

\begin{table*}[ht]
\centering
\caption{The drift detection ratio of ablation test. The highlighted values are greater than the baseline, which means the component increased the sensitivity of drift detection.}
\label{tab:3}
\begin{tabular}{llllllllllll}
\toprule
Drift Severity Level & 0     & 1     & 2     & 3     & 4     & 5     & 6     & 7     & 8     & 9     & 10    \\
\midrule
1. Complete             & 0.012 & 0.011 & 0.021 & 0.067 & 0.150 & 0.288 & 0.473 & 0.636 & 0.888 & 0.916 & 0.919 \\
2. MV Impute         & \cellcolor{blue!25}0.026 & \cellcolor{blue!25}0.027 & \cellcolor{blue!25}0.034 & \cellcolor{blue!25}0.075 & \cellcolor{blue!25}0.174 & \cellcolor{blue!25}0.296 & 0.429 & 0.524 & 0.656 & 0.684 & 0.713 \\
3. MDL               & \cellcolor{blue!25}0.034 & \cellcolor{blue!25}0.032 & \cellcolor{blue!25}0.061 & \cellcolor{blue!25}0.101 & \cellcolor{blue!25}0.293 & \cellcolor{blue!25}0.411 & \cellcolor{blue!25}0.585 & \cellcolor{blue!25}0.739 & \cellcolor{blue!25}0.951 & \cellcolor{blue!25}0.988 & \cellcolor{blue!25}0.995 \\
4.1 MDL-FWF-Gau      & 0.000 & 0.000 & 0.000 & 0.000 & 0.001 & 0.006 & 0.022 & 0.114 & 0.285 & 0.465 & 0.609 \\
4.2 MDL-FWF-Tri      & 0.000 & 0.000 & 0.000 & 0.001 & 0.004 & 0.007 & 0.026 & 0.078 & 0.194 & 0.364 & 0.542 \\
5.1 MDL-FWF-Gau-TopQ & 0.000 & 0.000 & 0.001 & 0.005 & 0.052 & 0.137 & 0.280 & 0.434 & 0.745 & 0.894 & \cellcolor{blue!25}0.953 \\
5.2 MDL-FWF-Tri-TopQ & 0.003 & 0.002 & 0.006 & 0.020 & 0.117 & 0.229 & 0.369 & 0.540 & 0.853 & \cellcolor{blue!25}0.956 & \cellcolor{blue!25}0.977 \\
\bottomrule
\end{tabular}
\end{table*}

\noindent\textbf{Findings and discussion}.

Fig. \ref{fig:8} summarizes the results of Experiment 2, and Table \ref{tab:3} records the average drift detection ratio in terms of the drift severity level. Further, the Pearson’s correlation coefficients for each configuration follows. Bold indicates the best configuration.
\begin{itemize}
    \item Config. 1 vs. 2: 0.798
    \item Config. 1 vs. 3: 0.922
    \item Config. 1 vs. 4.1: 0.651
    \item Config. 1 vs. 4.2: 0.694
    \item \textbf{Config. 1 vs. 5.1: 0.924}
    \item \textbf{Config. 1 vs. 5.2: 0.944}
\end{itemize}
These coefficients reveal the impact of each configuration on the drift detection results. Noting that the closer the coefficient to 1, the better the missing value handling method is.

According to the coefficient, it is clear that simply applying data imputation then concept drift detection (Config. 2, MV Impute) did not produce very accurate results. Table \ref{tab:3} shows that imputation might increase sensitivity for drifts of low severity but, with very severity drifts, it could have a opsite impact. Config. 3, where the distance between observations was estimated with masked distance learning, had a coefficient (0.922) very close to the baseline. However, the average drift detection ratios shown in Table \ref{tab:3} tell us that MDL increased the drift sensitivity in most cases. This opens the possibility that MDL may have a high false alarm when there is no drift and, therefore, will not reach the required confidence level. By contrast, Config. 4.1 and 4.2 (MDL-FWF with Gaussian and triangular functions) performed very poorly. These configurations treat distance estimation errors as an uncertainty issue and apply a fuzzy membership function to model the weight of observations. However, in some cases, the distance estimation errors can be very large, which results in an overestimation of the observations’ weights, in turn reducing the sensitivity of drift detection. Our solution was to choose the top-Q clusters only to handle these overestimated weights. As Configs. 5.1 and 5.2 show, performance with this approach was almost as good as the baseline without increasing the false alarm rate, where the baseline was the drift detection on the complete data sets.

\begin{figure*}[ht]
    \centering 

    \subfloat[Average Result on Complete Data Set]{\includegraphics[width=0.35\textwidth]{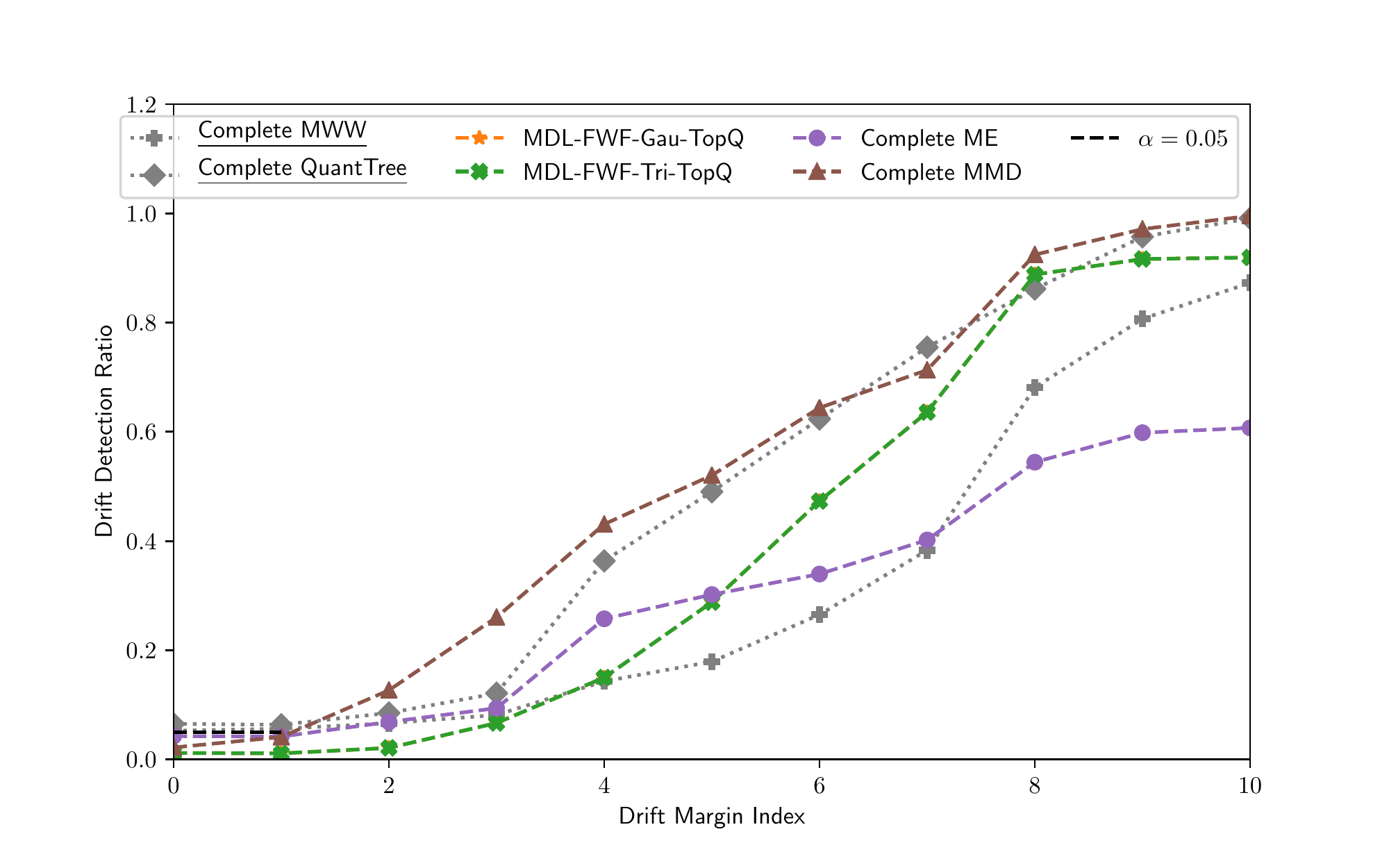}}
    \subfloat[Average Result on MV Data Set with Imputation]{\includegraphics[width=0.35\textwidth]{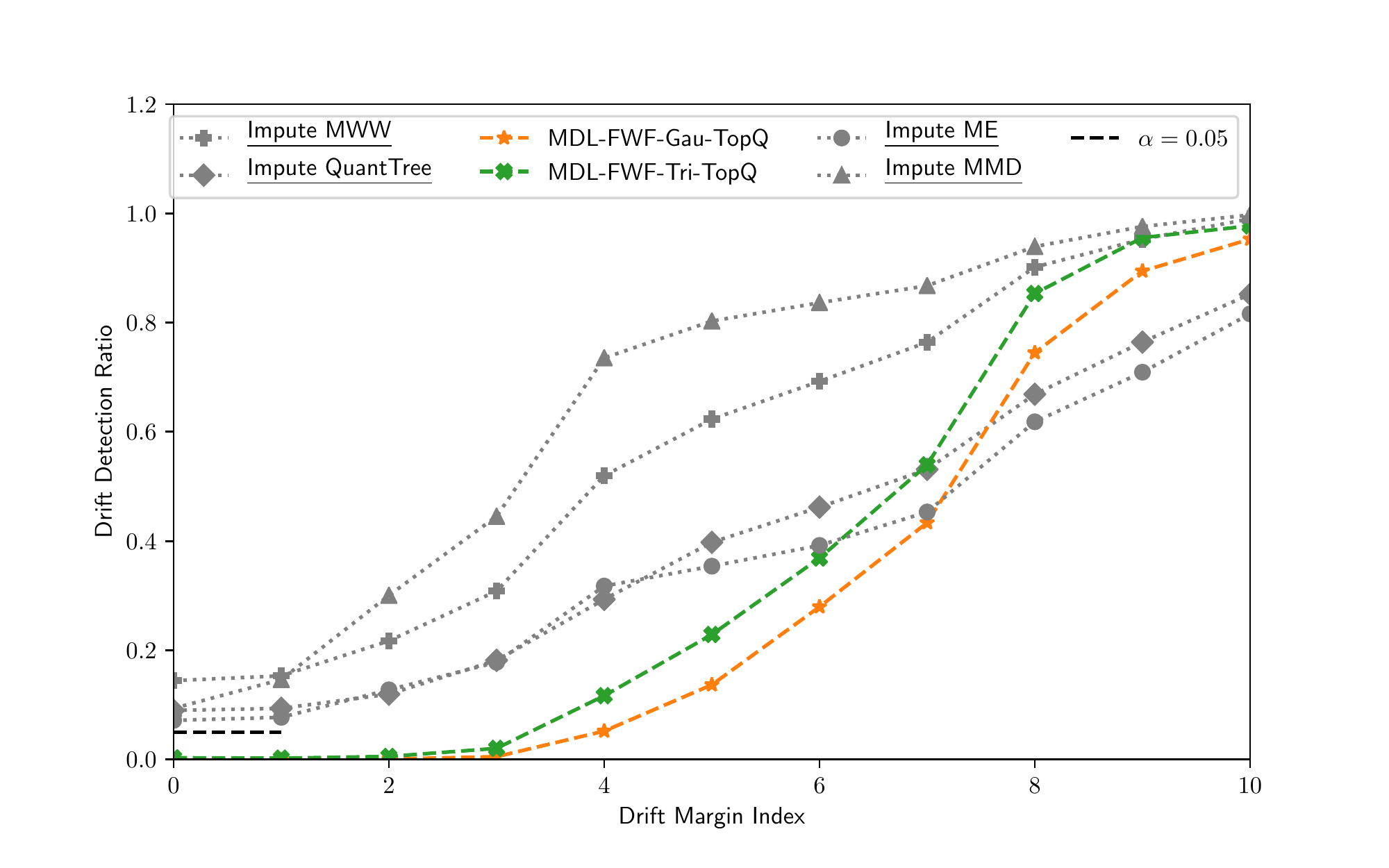}}

    \caption{The average drift detection ratio of different drift detection algorithms on synthetic data sets. The left subfigure shows the drift detection ratio on the complete data sets, that is, there are no missing values in the data sets. The right subfigure shows the drift detection ratio with the missing values imputed. The data sets used in the figures on the right-hand-side were generated by randomly removing values from the data sets used on the left-hand-side. (So the non-missing values in both data sets are the same.) These figures clearly show that the drift detection ratio of all compared algorithms has increased significantly. The drift detection ratios for the MWW, QuantTree, ME and MMD tests were around the desired threshold ($\alpha=0.05$) at drift margin index $0$, but the ratios far exceeded the threshold when the missing values appear.}
    \label{fig:9}
\end{figure*}

\subsection{A Comparative Study Against Other Drift Detection Algorithms}
\label{ss:exp3}
\noindent\textbf{Experiment 3 – A comparative study }

These experiments were designed to evaluate MDL-FWF’s performance in comparison to other concept drift detection algorithms. We chose four state-of-the-arts. The evaluation data sets and metric are the same as Experiment 2.

\noindent\textbf{Configurations}.

Multivariate Wald-Wolfowitz test (MWW test). This test was developed by Friedman and Rafsky \cite{MWWtest} as a multivariate generalization of the Wald-Wolfowitz test. This is the baseline for this experiment. The implementation is publicly available online\footnote{https://gist.github.com/vmonaco/e9ff0ac61fcb3b1b60ba}. No specific parameters are required to run this test.

QuantTree with Pearson statistics is a histogram-based change detection method for multivariate data streams \cite{Boracchi:QTree}. 
Pearson’s statistic is selected because it has the best reported performance. 
The implementation is available online\footnote{http://home.deib.polimi.it/boracchi/Projects/projects.html}. The partition size was set to the sample size divided by $50$, which is the same as the default setting for MDL-FWF.

MDL-FWF-TopQ Our proposed algorithm, again, implemented with two different membership functions, i.e., Gaussian and Triangle. The implementation code is available online\footnote{https://github.com/Anjin-Liu/TFS-MDL-FWF}.

ME Test is a generic normalized mean embedding (ME) test using a specified kernel. 
This is used in \cite{NIPS2015_5685, Arthur:ME2}. The implementation of the ME test is available online\footnote{https://github.com/wittawatj/interpretable-test}. The algorithm parameters were set to the defaults, except for $J$, which was set to $5$ following the authors' recommendation.

Quadratic MMD Test \cite{Arthur:QMMD}, where the null distribution is computed by permutation. 
The implementation is available online\footnote{https://github.com/wittawatj/interpretable-test}. 
The parameters were set following the authors' recommendation.

\noindent\textbf{Findings and discussion}.

\begin{table*}[ht]
\centering
\caption{The drift detection ratio on complete data sets. The underline indicates that the Type-I error exceeds the desired threshold $\alpha=0.05$.}
\label{tab:4}
\begin{tabular}{llllllllllll}
\toprule
Drift Severity Level & 0     & 1     & 2     & 3     & 4      & 5     & 6     & 7     & 8     & 9     & 10    \\
\midrule
Complete MWW         & \underline{0.053} & 0.056 & 0.066 & 0.082 & 0.143  & 0.179 & 0.265 & 0.383 & 0.681 & 0.807 & 0.873 \\
Complete QuantTree   & \underline{0.065} & 0.064 & 0.085 & 0.121 & 0.364  & 0.49  & 0.624 & 0.755 & 0.862 & 0.957 & 0.991 \\
Complete MDL   & 0.012 & 0.011 & 0.021 & 0.067 & 0.150  & 0.288 & 0.473 & 0.636 & 0.888 & 0.916 & 0.919 \\
MDL-FWF-Gau-TopQ     & 0.012 & 0.011 & 0.021 & 0.067 & 0.150  & 0.288 & 0.473 & 0.636 & 0.888 & 0.916 & 0.919 \\
MDL-FWF-Tri-TopQ     & 0.012 & 0.011 & 0.021 & 0.067 & 0.150  & 0.288 & 0.473 & 0.636 & 0.888 & 0.916 & 0.919 \\
Complete ME          & 0.043 & 0.042 & 0.069 & 0.094 & 0.2576 & 0.302 & 0.340 & 0.402 & 0.544 & 0.598 & 0.607 \\
Complete MMD         & 0.022 & 0.041 & 0.127 & 0.26  & 0.4308 & 0.520 & 0.644 & 0.713 & 0.924 & 0.971 & 0.996 \\
\bottomrule
\end{tabular}
\end{table*}

\begin{table*}[ht]
\centering
\caption{The drift detection ratio on data sets with missing values. 
The highlighted values are those with a higher value than the complete data set results.}
\label{tab:5}
\begin{tabular}{llllllllllll}
\toprule
Drift Severity Level & 0     & 1     & 2     & 3     & 4     & 5     & 6     & 7     & 8     & 9     & 10    \\
\midrule
Impute MWW           & \cellcolor{blue!25}\underline{0.144} & \cellcolor{blue!25}0.153 & \cellcolor{blue!25}0.217 & \cellcolor{blue!25}0.308 & \cellcolor{blue!25}0.520 & \cellcolor{blue!25}0.623 & \cellcolor{blue!25}0.692 & \cellcolor{blue!25}0.764 & \cellcolor{blue!25}0.902 & \cellcolor{blue!25}0.953 & \cellcolor{blue!25}0.990 \\
Impute QuantTree     & \cellcolor{blue!25}\underline{0.090} & \cellcolor{blue!25}0.094 & \cellcolor{blue!25}0.120 & \cellcolor{blue!25}0.182 & 0.293 & 0.398 & 0.462 & 0.531 & 0.669 & 0.764 & 0.852 \\
Impute MDL     & \cellcolor{blue!25}0.026 & \cellcolor{blue!25}0.027 & \cellcolor{blue!25}0.034 & \cellcolor{blue!25}0.075 & \cellcolor{blue!25}0.174 & \cellcolor{blue!25}0.296 & 0.429 & 0.524 & 0.656 & 0.684 & 0.713 \\
MDL-FWF-Gau-TopQ     & 0.000 & 0.000 & 0.001 & 0.005 & 0.052 & 0.137 & 0.280 & 0.434 & 0.745 & 0.894 & \cellcolor{blue!25}0.953 \\
MDL-FWF-Tri-TopQ     & 0.003 & 0.002 & 0.006 & 0.020 & 0.117 & 0.229 & 0.369 & 0.540 & 0.853 & \cellcolor{blue!25}0.956 & \cellcolor{blue!25}0.977 \\
Impute ME            & \cellcolor{blue!25}\underline{0.072} & \cellcolor{blue!25}0.077 & \cellcolor{blue!25}0.128 & \cellcolor{blue!25}0.178 & \cellcolor{blue!25}0.318 & \cellcolor{blue!25}0.354 & \cellcolor{blue!25}0.392 & \cellcolor{blue!25}0.453 & \cellcolor{blue!25}0.618 & \cellcolor{blue!25}0.709 & \cellcolor{blue!25}0.816 \\
Impute MMD           & \cellcolor{blue!25}\underline{0.094} & \cellcolor{blue!25}0.146 & \cellcolor{blue!25}0.300 & \cellcolor{blue!25}0.445 & \cellcolor{blue!25}0.735 & \cellcolor{blue!25}0.803 & \cellcolor{blue!25}0.836 & \cellcolor{blue!25}0.868 & \cellcolor{blue!25}0.939 & \cellcolor{blue!25}0.976 & \cellcolor{blue!25}0.997
\\ \bottomrule
\end{tabular}
\end{table*}

The drift detection results are plotted in Fig. \ref{fig:9}, and the average ratios on each of the data sets are shown in Tables \ref{tab:4} and \ref{tab:5}. The Pearson’s correlation coefficients of the 50 drift detection ratios were as follows:
\begin{itemize}
    \item ME missing vs. complete 0.235
    \item MWW missing vs. complete 0.667
    \item QuantTree missing vs. complete 0.785
    \item MMD missing vs. complete 0.887
    \item \textbf{MDL-FWF-Gau-TopQ vs. complete: 0.924}
    \item \textbf{MDL-FWF-Tri-TopQ vs. complete: 0.944}
\end{itemize}

Missing values impacted the ME test the most, followed by the MWW test, QuantTree, MMD test and the MDL-FWF methods. These results confirm our assumption that missing values introduce uncertainty into distribution-based concept drift detection. Although it is difficult to examine their general performance on all distributions, these results, at least, demonstrate the need to consider missing values when dealing with distribution changes. Another conclusion we draw from these results is that fuzzy theory has good potential for handle missing values in tasks that required accurate and sensitive concept drift detection.

\subsection{Drift detection on real-world data sets with synthetic drifts and missing values}
\label{ss:exp4}
\noindent\textbf{Experiment 4 – Evaluation with real-world data}

Our final test was to compare MLD-FWF with state-of-the-art algorithms in real-world scenarios. 

\begin{table*}[ht]
\centering
\caption{Drift detection Type-I errors with the real-world data sets. TtDiff is the sum of absolute difference between the Type-I errors with and without missing values. Impu stands for iteratively imputed missing values. Comp is the complete data set with no missing values. Rank is the sort order in terms of AgDiff.}
\label{tab:9}
\begin{tabular}{lllllllllllllll}
\toprule
      & \multicolumn{2}{l}{Real-I.a} & \multicolumn{2}{l}{Real-I.b} & \multicolumn{2}{l}{Real-II.a} & \multicolumn{2}{l}{Real-II.b} & \multicolumn{2}{l}{Real-III.a} & \multicolumn{2}{l}{Real-III.b} & \multirow{2}{*}{TtDiff} & \multirow{2}{*}{Rank} \\
      & Impu          & Comp         & Impu          & Comp         & Impu          & Comp          & Impu          & Comp          & Impu           & Comp          & Impu           & Comp          &                         &                       \\
\toprule
MWW   & \underline{0.508}         & 0.008        & \underline{0.874}         & \underline{0.076}        & 0.038         & 0.006         & 0.040         & 0.014         & 0.014          & 0.002         & 0.028          & 0.004         & 1.39                    & 7                     \\
QuantTree & 0.048         & 0.000        & \underline{0.170}         & 0.036        & \underline{0.068}         & 0.022         & \underline{0.200}         & \underline{0.092}         & \underline{0.278}          & \underline{0.062}         & \underline{0.482}          & \underline{0.090}         & 0.94                    & 6                     \\
MDL & 0.002         & 0.014        & 0.016         & 0.004        & 0.010         & 0.044         & 0.000         & 0.000         & 0.014          & 0.028         & 0.038          & 0.008         & 0.10                    & 2                     \\
MDL-FWF-Gau-TopQ   & 0.000         & 0.014        & 0.012         & 0.004        & 0.010         & 0.044         & 0.000         & 0.000         & 0.014          & 0.028         & 0.000          & 0.008         & 0.08                    & 1                     \\
MDL-FWF-Tri-TopQ   & 0.000         & 0.014        & 0.022         & 0.004        & 0.002         & 0.044         & 0.000         & 0.000         & 0.000          & 0.028         & 0.000          & 0.008         & 0.11                    & 3                     \\
ME    & \underline{0.164}         & 0.020        & 0.040         & \underline{0.164}        & \underline{0.052}         & \underline{0.080}         & 0.004         & 0.000         & \underline{0.092}          & \underline{0.064}         & \underline{0.186}          & 0.010         & 0.50                    & 5                     \\
MMD   & \underline{0.226}         & 0.020        & \underline{0.256}         & \underline{0.060}        & 0.008         & 0.008         & 0.002         & 0.000         & 0.000          & 0.000         & 0.012          & 0.010         & 0.41                   & 4                                       
\\ \bottomrule
\end{tabular}
\end{table*}

\begin{table*}[ht]
\centering
\caption{Drift detection Type-II errors with the real-world data sets.}
\label{tab:10}
\begin{tabular}{lllllllllllllll}
\toprule
      & \multicolumn{2}{l}{Real-I.a} & \multicolumn{2}{l}{Real-I.b} & \multicolumn{2}{l}{Real-II.a} & \multicolumn{2}{l}{Real-II.b} & \multicolumn{2}{l}{Real-III.a} & \multicolumn{2}{l}{Real-III.b} & \multirow{2}{*}{TtDiff} & \multirow{2}{*}{Rank} \\
      & Impu          & Comp         & Impu          & Comp         & Impu          & Comp          & Impu          & Comp          & Impu           & Comp          & Impu           & Comp          &                         &                       \\
\toprule
MWW   & 0.024         & 0.824        & 0.013         & 0.593        & 0.909         & 0.933         & 0.843         & 0.960         & 0.985          & 1.005         & 0.966          & 0.995         & 1.57                    & 6                     \\
QuantTree & 0.294         & 0.785        & 0.369         & 0.459        & 0.733         & 0.970         & 0.245         & 0.987         & 0.505          & 0.261         & 0.431          & 0.829         & 2.20                    & 7                     \\
MDL & 0.808         & 0.470        & 0.780         & 0.868        & 0.415         & 0.426         & 0.308         & 0.535         & 0.963          & 0.967         & 0.634          & 0.816         & 0.85                    & 3                     \\
MDL-FWF-Gau-TopQ   & 0.500         & 0.469        & 0.996         & 0.862        & 0.483         & 0.423         & 0.757         & 0.532         & 0.960          & 0.969         & 0.981          & 0.817         & 0.62                    & 1                     \\
MDL-FWF-Tri-TopQ   & 0.481         & 0.468        & 0.971         & 0.865        & 0.418         & 0.423         & 1.002         & 0.536         & 1.006          & 0.965         & 1.009          & 0.816         & 0.82                    & 2                     \\
ME    & 0.729         & 0.881        & 0.918         & 0.786        & 0.796         & 0.932         & 0.970         & 0.927         & 0.972          & 0.703         & 0.467          & 0.777         & 1.04                    & 5                     \\
MMD   & 0.168         & 0.544        & 0.133         & 0.323        & 0.142         & 0.188         & 0.226         & 0.367         & 1.008          & 1.009         & 0.500          & 0.626         & 0.88                    & 4         
\\ \bottomrule
\end{tabular}
\end{table*}

\noindent\textbf{Data sets}.


HIGGS Bosons and Background Data Set\cite{Gretton:MMD}. The objective of this data set is to distinguish signatures of the processes that produce Higgs boson particles from background processes that do not. We selected four low-level indicators of azimuthal angular momenta for four particle jets as features as the same as \cite{Gretton:MMD}. The jet momenta distributions of the background processes are denoted as Back ($\mathcal{B}$), and the processes that produce Higgs bosons are denoted as Higgs ($\mathcal{A}$). 
There were four types of data integration: Back-Back (Type-I Real I.a), where both sample sets were drawn from Back; Higgs-Higgs (Type-I Real I.b), where both sample sets were drawn from Higgs; Back-Higgs (Type-II Real I.a); and Higgs-Back (Type-II Real I.b)

Arabic Digit Mixture Data Set \cite{FastKSTest}. This data set contains audio features of 88 people (44 females and 44 males) pronouncing Arabic digits between $0$ and $9$. We applied the same configuration as Denis et al. \cite{FastKSTest}. The data set has $26$ attributes which are a replacement mean and standard deviation for $13$ time series. 
Mixture distribution $\mathcal{A}$ contained randomly selected samples of both males and females, with female labels from $0$ to $4$ and male labels from $5$ to $9$. Mixture distribution $\mathcal{B}$ reversed the labels, i.e., males from $0$ to $4$ and females from $5$ to $9$. The drift detection on $\mathcal{A}-\mathcal{A}$ is Type-I Real II.a; $\mathcal{B}-\mathcal{B}$ is Type-I Real II.b; $\mathcal{A}-\mathcal{B}$ is Type-II Real II.a; and $\mathcal{B}-\mathcal{A}$ is Type-II Real II.b.

Insects Mixture Data Set \cite{FastKSTest}. This data set contains 49 features from a laser sensor. The task is to distinguish between 5 possible specimens of flying insects that pass through a laser in a controlled environment (Flies, Aedes, Tarsalis, Quinx, and Fruit). A preliminary analysis showed no drift in the feature space. However, the class distributions gradually change over time. To simulate drift in multiple clusters, we selected the samples from different insects and grouped them together at different percentages; thus, the data distribution varies. The populations are generated baBsed on the ratios that $\mathcal{A}$=$\{Flies:0.2, Aedes:0.2, Tarsalis:0.2, Quinx:0.2, Fruit:0.2\}$ and $\mathcal{
B}$$=$$\{Flies:0.14, Aedes:0.14, Tarsalis:0.2, Quinx:0.2, Fruit:0.32\}$

\noindent\textbf{Evaluation metric}.

According to David \cite{David:SmallSample}, at least $25$ samples are required to estimate Pearson, Kendall, and Spearman correlations. Therefore, the results were evaluated based on the absolute total difference between the complete and the imputed drift detection ratio in terms of Type-I and Type-II errors, i.e., 
\begin{equation*}
    \mathrm{TtDiff} = \sum_{i}^{\#dr}|dr^{comp}_i - dr^{impu}_i|,
\end{equation*}
where $dr^{comp}_i$ indicates the drift detection ratio on complete data sets, $dr^{impu}_i$ is on missing values with imputation, and $\#dr$ denotes the number of comparisons.

\noindent\textbf{Findings and discussion}.

The results of Experiment 4 are summarized in Tables \ref{tab:9} and \ref{tab:10}. The results show a similar pattern to Experiment 3, that is, imputing missing values introduces uncertainty and decreases the sensitivity of drift detection. However, MDL-FWF is more robust to missing values than its comparators. \hly{From Tables  \ref{tab:9} and \ref{tab:10}, we can see that MMD was the most powerful drift detector on the complete data set, but, after introducing missing values, the Type-I errors increased dramatically and the Type-II errors dropped significantly. The total difference in Type-I errors for MMD on the six data sets was $0.41$, and the difference in Type-II errors was $0.88$. Thus the mean of all errors was $\frac{0.41+0.88}{6+6}=0.1075$. In contrast, MDL-FWF had a mean of only $0.0583$ for all errors. This is an improvement of nearly 46\%.}
Such a change in the drift detection results is dangerous because it may increase the number of false alarms and result in unnecessary warnings. 
Based on the results of these four experiments, we conclude that missing values should be carefully addressed when comparing sample distributions and we find that using fuzzy set theory to model the uncertainty introduced by missing values has a positive effect on distribution discrepancy analysis.

\section{Conclusion}
\label{s:V}

\hlr{In this paper, we investigated the influence of missing values on concept drift detection and imputation from a task perspective rather than a value perspective - that is, estimating the statistics we need, as opposed to estimating the missing values then calculating the statistics. The proposed algorithm is one integrated concept drift detection algorithm which can handle missing values, but not an imputation algorithm combined with a concept drift detection algorithm.}
The results show that applying data imputation then conducting drift detection impairs the performance of today's drift detection algorithms. We attribute this phenomenon to imputation errors and propose an observation fuzzification model to account for the errors. Our proposed drift detection algorithm consists of two major steps. The first is a masked distance learning (MDL) scheme that estimates the distance between observations with missing values and its surrounding cluster centroids. The second is the fuzzy-weighted frequency (FWF) method that considers MDL's estimation errors and uses fuzzy membership to normalize the weighted frequency of observations. Four experiments were conducted to show the efficacy of our method.

In future work, we need to explore a method that can address the disadvantages of complete case analysis, as well as a robust histogram construction method that can handle missing values in extreme cases. \hlc{The dimensionality of the data is also worth mentioning. Since two observations are concatenated horizontally, which doubles the number of dimensions, computation costs can increase dramatically with high-dimensional data sets. We consider this as our next challenge to solve. Our intention is to begin by exploring whether any dimension reduction techniques might mitigate the problem.}\hlr{At the current stage, we have only applied a very shallow concept of fuzzy-distance with our fuzzy-distance estimation. However, there is still potential for a distribution estimation based on uncertainty measures. This is a more theoretically-oriented strategy that we intend to explore in a future study}

\section*{Acknowledgments}
The work presented in this paper was supported by the
Australian Research Council (ARC) under Discovery Project DP190101733.

\ifCLASSOPTIONcaptionsoff
  \newpage
\fi



\bibliographystyle{jabbrv_ieeetr}
\bibliography{bib/reference}

%



%


\begin{IEEEbiography}[{\includegraphics[width=1in,height=1.25in, clip]{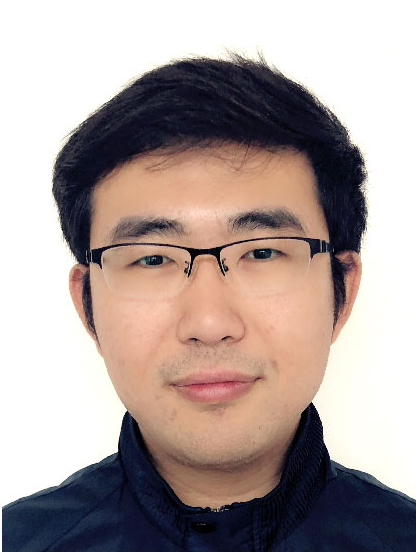}}]{Anjin Liu}
is a Postdoctoral Research Associate in the A/DRsch Centre for Artificial Intelligence, Faculty of Engineering and Information Technology, University of Technology Sydney, Australia. He received the BIT degree (Honour) at the University of Sydney in 2012. His research interests include concept drift detection, adaptive data stream learning, multi-stream learning, machine learning and big data analytics.
\end{IEEEbiography}

\begin{IEEEbiography}[{\includegraphics[width=1in,height=1.25in,clip,keepaspectratio]{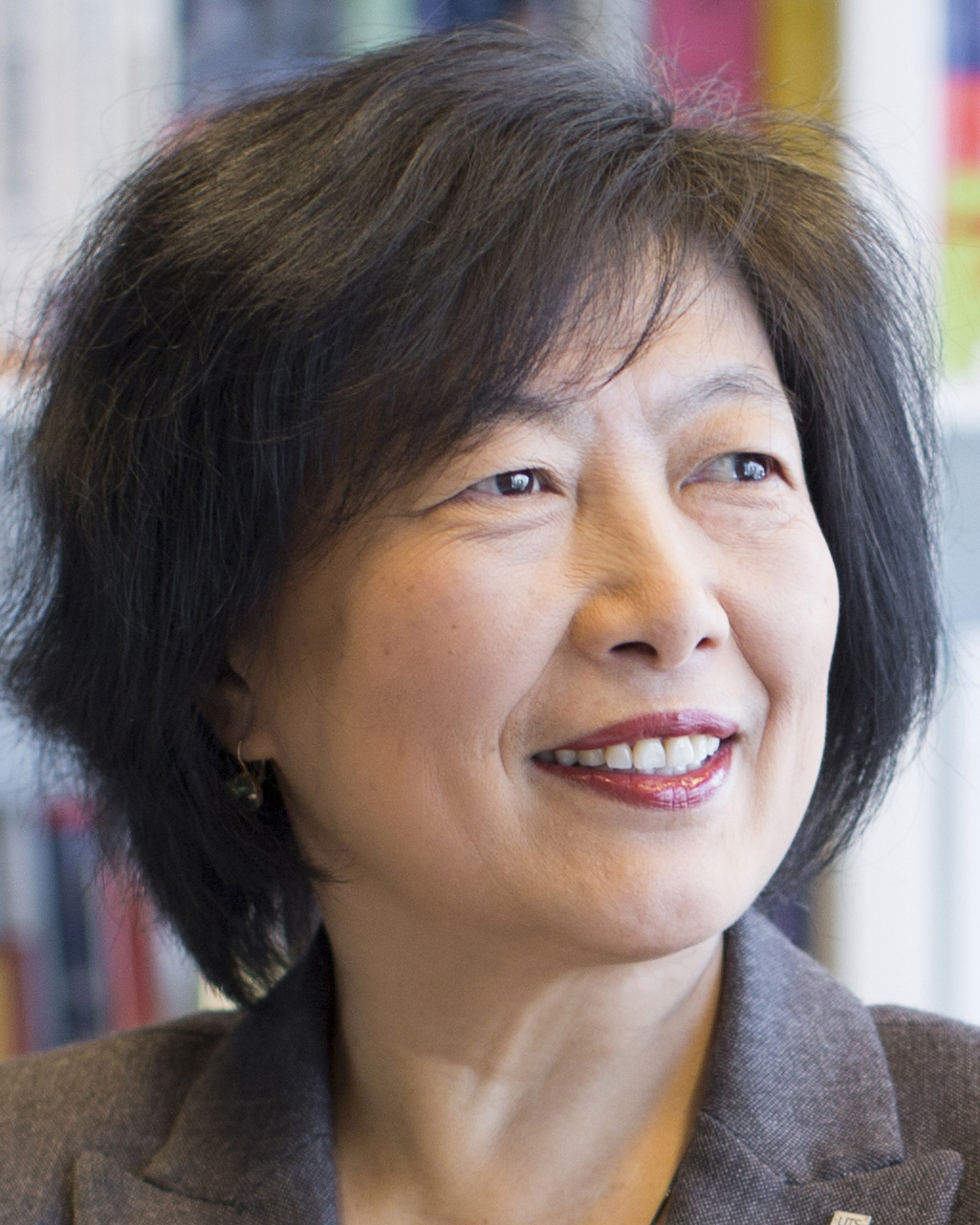}}]{Jie Lu}(F'18) is an Australian Laureate Fellow, IEEE Fellow and IFSA Fellow. She is an internationally renowned scientist in the areas of computational intelligence, specifically in fuzzy transfer learning, concept drift, decision support systems, and recommender systems. She is the Director of the Centre for Artificial Intelligence (CAI), which has over 200 staff and students working on over 50 research projects, and the Associate Dean (Research Excellence) in the Faculty of Engineering and Information Technology at the University of Technology Sydney (UTS).

She has published six research books and over 450 papers in Artificial Intelligence, IEEE Transactions on Fuzzy Systems, IEEE Transactions on Neural Networks \& Learning Systems, Decision Support Systems and other refereed journals and conference proceedings. She has won 10 ARC Discovery projects and led 15 industry projects. She has supervised 40 PhD students to completion. She serves as Editor-In-Chief for Knowledge-Based Systems (Elsevier) and Editor- In-Chief for International Journal on Computational Intelligence Systems (Atlantis). She is the Chair of IEEE CIS Emerging Technologies Technical Committee and the Chair of IEEE CIS Subcommittee for Senior Members.She has delivered 25 keynote speeches at IEEE and other international conferences and chaired 15 international conferences. She has received the UTS Medal for Research and Teaching Integration (2010),UTS Medal for research excellence (2019), finalist of the Australian Eureka Prize for Advanced Data Science (2017, 2018), the Computer Journal Wilkes Award (2018), the IEEE Transactions on Fuzzy Systems Outstanding Paper Award (2019), the Australian Most Innovative Engineer Award (2019) and other national and international awards.

\end{IEEEbiography}

\begin{IEEEbiography}[{\includegraphics[width=1in,height=1.25in, clip]{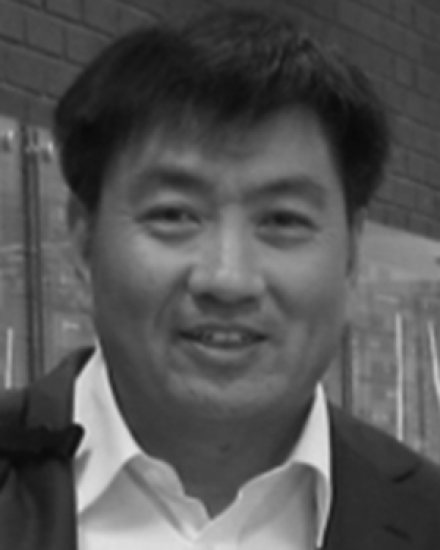}}]{Guangquan Zhang}
is an Associate Professor and Director of the Decision Systems and e-Service Intelligent (DeSI) Research Laboratory at the University of Technology Sydney, Australia. He received the Ph.D. degree in applied mathematics from Curtin University of Technology, Australia, in 2001.

His research interests include fuzzy machine learning, fuzzy optimization, and machine learning. He has authored five monographs, five textbooks, and  460 papers including 220 refereed international journal papers
Dr. Zhang has won seven Australian Research Council (ARC) Discovery Projects grants and many other research grants. He was awarded an ARC QEII fellowship in 2005.
He has served as a member of the editorial boards of several international journals, as a guest editor of eight special issues for IEEE transactions and other international journals, and co-chaired several international conferences and workshops in the area of fuzzy decision-making and knowledge engineering.
\end{IEEEbiography}




\end{document}